\documentclass[12pt]{article}
\usepackage{siunitx}
\usepackage{hyperref}
\hypersetup{hidelinks}

\usepackage{amsmath, 
	amsfonts, 
	amssymb, 
	bm,
	mathtools,
	amsthm}
\usepackage{graphicx,
	}
\usepackage{booktabs}
\usepackage{multirow}
\usepackage{pgfplots,
	pgfplotstable,
	tikz,
	eso-pic}
\usetikzlibrary{patterns,
	plotmarks}
\pgfplotsset{compat=1.12}

\usepackage[english,
	onelanguage,
	ruled,
	vlined,
	longend,
	titlenotnumbered]{algorithm2e}
	
\usepackage[nospace]{cite}

\usepackage{blindtext,
	ifpdf}
\usepackage{setspace}
\usepackage{paralist}

\usepackage{filecontents}

\newcommand{\bfA}{\mathbf{A}}
\newcommand{\bfb}{\mathbf{b}}
\newcommand{\hbfb}{\hat{\mathbf{b}}}

\newcommand{\bfe}{\mathbf{e}}

\newcommand{\bfF}{\mathbf{F}}
\newcommand{\bfG}{\mathbf{G}}

\newcommand{\bfg}{\mathbf{g}}
\newcommand{\bfH}{\mathbf{H}}
\newcommand{\bfI}{\mathbf{I}}
\newcommand{\bfJ}{\mathbf{J}}

\newcommand{\bfK}{\mathbf{K}}

\newcommand{\bfM}{\mathbf{M}}
\newcommand{\bfN}{\mathbf{N}}
\newcommand{\calN}{\mathcal{N}}
\newcommand{\calC}{\mathcal{C}}

\newcommand{\bfn}{\mathbf{n}}
\newcommand{\calO}{\mathcal{O}}
\newcommand{\bfp}{\mathbf{p}}
\newcommand{\hbfp}{\hat{\mathbf{p}}}

\newcommand{\bfP}{\mathbf{P}}

\newcommand{\bfQ}{\mathbf{Q}}

\newcommand{\bfR}{\mathbf{R}}

\newcommand{\hbfR}{\hat{\bfR}}
\newcommand{\bbR}{\mathbb{R}}

\newcommand{\bfS}{\mathbf{S}}

\newcommand{\bfu}{\mathbf{u}}
\newcommand{\bfv}{\mathbf{v}}
\newcommand{\hbfv}{\hat{\bfv}}

\newcommand{\bfw}{\mathbf{w}}

\newcommand{\bfx}{\mathbf{x}}

\newcommand{\hbfx}{\hat{\mathbf{x}}}
\newcommand{\calX}{\mathcal{X}}

\newcommand{\bfy}{\mathbf{y}}

\newcommand{\boomega}{\boldsymbol{\omega}}

\newcommand{\bfzero}{\mathbf{0}}

\DeclareMathOperator{\diag}{diag}

\DeclareMathOperator{\Ker}{Ker}

\newtheoremstyle{mystyle}
{}
{}
{\itshape}
{}
{\bfseries}
{.}
{ }
{}

\newtheorem{mydef}{Definition}
\newtheorem{mytheorem}{Proposition}
\newtheorem{mytheorem2}{Theorem}
\newtheorem{example}{Example}

\newtheorem{ass}{Assumption}
\usepackage[table]{colortbl}
\usepackage{xcolor}

\pgfplotstableread[col sep=comma]{rmseExp.dat}\rmseExp
\usepackage{mleftright}

\usepackage{listings}
\usepackage{xcolor}

\lstdefinelanguage{BibTeX}
{keywords={%
		@article,@book,@collectedbook,@conference,@electronic,@ieeetranbstctl,%
		@inbook,@incollectedbook,@incollection,@injournal,@inproceedings,%
		@manual,@mastersthesis,@misc,@patent,@periodical,@phdthesis,@preamble,%
		@proceedings,@standard,@string,@techreport,@unpublished%
	},
	comment=[l][\itshape]{@comment},
	sensitive=false,
}

\begin{document}

{\center{This paper has been accepted in:

\vspace{1 cm} 

\emph{IEEE Sensors Journal} Vol. 19(4), 2019. 

\vspace{1 cm}
	DOI  : 10.1109/JSEN.2018.2882714
	
	\vspace{1 cm}
	}}
	The present online version posted by the authors is augmented with an extended appendix which provides  {much} more details to practitioners and is not  explicitly displayed in the journal paper. Notably, all the pseudo algorithms that are used in the simulations and experiments are provided in detail (whereas, owing to space limits, the journal paper gives only their general form and one has to work out the details). Nevertheless, please cite the \emph{IEEE Sensors Journal} paper when referring to the present paper as follow: 
	
	\begin{lstlisting}[language=BibTeX]
	@article{brossardExploiting2019,
	author={Martin {Brossard} and Axel {Barrau} and Silv\`ere {Bonnabel}},
	journal={IEEE Sensors Journal},
	title={Exploiting Symmetries to Design EKFs With Consistency Properties for Navigation and SLAM},
	year={2019},
	volume={19},
	number={4},
	pages={1572-1579},
	doi={10.1109/JSEN.2018.2882714},
	ISSN={1530-437X},
	month={Feb},
	}
	\end{lstlisting}
		\vspace{\fill}

	IEEE copyright applies. Personal use of this material is permitted.  Permission from IEEE must be obtained for all other uses, in any current or future media, including reprinting/republishing this material for advertising or promotional purposes, creating new collective works, for resale or redistribution to servers or lists, or reuse of any copyrighted component of this work in other works.

	\title{Exploiting Symmetries to Design EKFs with Consistency Properties for Navigation and SLAM}
%
%

	\author{Martin \textsc{Brossard}\thanks{M. Brossard and S. Bonnabel are with MINES ParisTech, PSL Research University, Centre for Robotics, 60 Boulevard Saint-Michel, 75006 Paris, France (email: martin.brossard@mines-paristech.fr;silvere.bonnabel@mines-paristech.fr).}	, \and Axel \textsc{Barrau}\thanks{A. Barrau is with Safran Tech, Groupe Safran, Rue des Jeunes Bois-Ch\^ateaufort, 78772, Magny Les Hameaux Cedex, France (email: axel.barrau@safrangroup.com).} \and and Silv\`ere \textsc{Bonnabel}\footnotemark[1]
	}

	\maketitle
	\begin{abstract}
		The Extended Kalman Filter (EKF) is both the historical algorithm for multi-sensor fusion and still state of the art in numerous industrial applications. However, it may prove inconsistent in the presence of   unobservability under a group of transformations. In this paper we first   build   an alternative EKF based on an alternative nonlinear state error. This EKF is intimately related to the theory of the Invariant EKF (IEKF). Then, under a  simple compatibility assumption between the error and the transformation group, we prove the linearized model of the alternative EKF  automatically captures the unobservable directions, and many desirable properties of the linear case then directly follow. This provides a novel fundamental result in  filtering theory. We apply the   theory to multi-sensor fusion for navigation, when all the sensors are attached to the vehicle and do not have access to absolute information,  as typically occurs in GPS-denied environments.  In the   context of Simultaneous Localization And Mapping (SLAM),  Monte-Carlo runs and comparisons to OC-EKF, robocentric EKF, and optimization-based smoothing algorithms (iSAM)  illustrate the results. The proposed EKF is also proved to outperform standard EKF and to achieve comparable performance to iSAM on a publicly available real dataset for multi-robot SLAM. 
	\end{abstract}
	

\section{Introduction}\label{sec:int}
Multi-sensor fusion for navigation of autonomous and non-autonomous  vehicles,  or for  Simultaneous Localization And Mapping (SLAM) or Visual Inertial Odometry (VIO) is classically handled by the Extended Kalman Filter (EKF). Although powerful alternative techniques have since emerged, the EKF is both  the historical algorithm -  originally implemented in the Apollo program - and still a prevalent algorithm in the academia and in  the industry, see e.g.,   \cite{cadena_past_2016,bloesch_iterated_2017,sun_robust_2018}.

One major limitation of the EKF   is its  inconsistency, that is, the filter returns a covariance matrix that is too optimistic \cite{bar-shalom_state_2002},  
leading to inaccurate estimates. EKF inconsistency in the context of SLAM  has been the object
of many papers, see e.g. \cite{julier_counter_2001,hesch_camera-imu-based_2014,li_high-precision_2013,castellanos_limits_2004,
	bailey_consistency_2006,
	huang_towards_2017,
	huang_analysis_2008,
	huang_towards_2014, huang_optimal-state-constraint_2018,huang_observability-based_2011}. Theoretical analysis   \cite{julier_counter_2001,
	huang_analysis_2008,huang_observability-based_2011} reveals inconsistency is caused by the inability of EKF to reflect the unobservable
degrees of freedom of SLAM. Indeed, the filter tends to erroneously acquire
information along the directions spanned by these unobservable degrees of freedom. The Observability Constrained (OC)-EKF \cite{huang_observability-based_2011,huang_towards_2017} constitutes one of the most advanced solutions to remedy this problem and has been fruitfully adapted, e.g. for VIO, cooperative localization, and unscented Kalman filter \cite{huang_observability-based_2010, li_high-precision_2013, huang_quadratic-complexity_2013}. The idea is to pick a
linearization point that is such that the unobservable subspace
``seen" by the filter is of appropriate dimension. 

In this paper we propose a novel general theory. We first propose to build EKFs based on an alternative error $\bfe=\eta(\bfx,\hat\bfx)$, generalizing the Invariant EKF (IEKF) methodology \cite{barrau_invariant_2017}. This means the covariance matrix $\bfP$ reflects the dispersion of $\bfe$, and not of $\bfx-\hat\bfx$. When unobservability stems from symmetries, the technique may resolve the consistency issues of the EKF. Indeed symmetries are encoded by the action $\phi_\alpha(\bfx)$ of a transformation group $G$ \cite{OLVER}, where $\alpha\in G$ denotes the corresponding infinitesimal unobservable transformation. Under the simple condition that \textbf{the image of matrix $\frac{\partial}{\partial\alpha}\eta(\phi_\alpha(\bfx),\bfx)$  is independent of $\bfx$}, the EKF based on $\bfe$ is proved to possess the desirables properties of the linear case regarding unobservability,  and is thus consistent. 
\subsection{Specific Application to SLAM}

The specific application   to SLAM was released by the authors on Arxiv in 2015  \cite{barrau_ekf-slam_2015}  and encountered immediate successes reported in \cite{zhang2017convergence, wu_invariant-ekf_2017,heo_consistent_2018,brossard2017unscented,caruso_2018,barrau_invariant_2018,heo2018consistent},  although the work was never published elsewhere. 

More precisely, we   noticed back in \cite{bonnabel_symmetries_2012} the SLAM problem bears a nontrivial Lie group structure. In the Arxiv preprint \cite{barrau_ekf-slam_2015}   we  formalized the group introduced in \cite{bonnabel_symmetries_2012} and called it $SE_{l+1}(3)$, and we proved that for odometry based SLAM, using the right invariant error of $SE_{l+1}(3)$ and devising an EKF based on this error, i.e., a Right-Invariant EKF (RIEKF), the linearized system possesses  the desirable properties of the linear case, since it automatically correctly captures unobservable directions for SLAM. 
Thus, virtually all properties of the linear Kalman filter regarding unobservability may be directly transposed: the information about unobservable  directions is non-increasing (see Proposition  \ref{mainthm}), the dimension of the unobservable subspace has appropriate dimension (this relates to the result of OC-EKF \cite{huang_quadratic-complexity_2013,huang_observability-based_2011, huang_observability-based_2010}), the filter's output is invariant to linear unobservable transformations, even if they are stochastic and thus change the EKF's covariance matrix along  unobservable directions  \cite{zhang2017convergence}. The right-invariant error for the proposed Lie group structure was also recently shown to lead to deterministic  observers  having  exponential  convergence properties in \cite{Mahony2017a}.

Along the same lines, using the right-invariant error of the group $SE_2(3)$ we proposed  in  \cite{barrau_invariant_2017} also led  to  alternative  consistent IEKF for visual inertial SLAM and VIO applications \cite{wu_invariant-ekf_2017, heo_consistent_2018,brossard2017unscented, caruso_2018, heo2018consistent}. In particular, \cite{heo_consistent_2018} demonstrates that an alternative Invariant MSCKF based on the right-invariant error of $SE_2(3)$  naturally
enforces the state vector to remain in the unobservable subspace, a consistency property which is preserved when considering point and line features \cite{heo2018consistent}, or when a network of magnetometers is  available \cite{caruso_2018}. 

\subsection{Paper's Organization}

Section \ref{gen:the} presents the general theory.  Section \ref{nav:ss}  applies the theory to the general problem of  navigation  in the  absence of absolute measurements, as typically occurs in GPS-denied environments.  Section \ref{sec:sim} is dedicated to SLAM  and  compares the proposed EKF to conventional EKF, OC-EKF \cite{huang_observability-based_2011}, robocentric mapping filter \cite{castellanos_robocentric_2007} and iSAM \cite{kaess_isam2:_2012, kummerle_g2o:_2011}. In Section  \ref{sec:exp},   we show our alternative EKF achieves comparable results to iSAM and outperforms the conventional EKF on a multi-robot SLAM experiment using  the UTIAS dataset \cite{leung_utias_2011}. 

Preliminary ideas and results can be found in the  2015 technical report \cite{barrau_ekf-slam_2015}. Although the present paper is a major rewrite, notably including a novel general theory encompassing  the particular application to SLAM of \cite{barrau_ekf-slam_2015}, and   novel comparisons and experiments,   \cite{barrau_ekf-slam_2015} serves as preliminary material  for the present paper. Matlab codes used for the paper are  available at \texttt{\url{https://github.com/CAOR-MINES-ParisTech/esde}}.

\section{General Theory}\label{gen:the}

Let us consider the following dynamical system in discrete time with state $\bfx_n \in \calX$ and observations $\bfy_n \in \bbR^p$:\begin{align}
\bfx_n &= f\left(\bfx_{n-1},\bfu_n,\bfw_n\right), \label{eq:dyn}\\
\bfy_n &= h\left(\bfx_n, \bfv_n\right), \label{eq:mes}
\end{align}
where $f(\cdot)$ is the function encoding the evolution of the system, $\bfw_n \sim \calN\left(\bfzero,\bfQ_n\right)$ is the Gaussian process noise,  $\bfu_n$ is the input, $h(\cdot)$ is the observation function and $\bfv_n \sim \calN\left(\bfzero,\bfR_n\right)$ a Gaussian measurement noise. 

We now define mathematical symmetries, see  \cite{OLVER,bonnabel_symmetry-preserving_2008}.

\begin{mydef}
	 An action of a (Lie) group $G$ on $\calX$ is defined as a family of bijective maps $\phi_\alpha:  \calX \rightarrow \calX,~\alpha\in G$ satisfying
	\begin{align}
	\forall \bfx \in\calX \quad &\phi_{Id}(\bfx)=\bfx, \label{eq:ga0}\\
	\forall \alpha,\beta\in G,\bfx \in\calX \quad &\phi_{\alpha}\left(\phi_{\beta}\left(\bfx\right)\right) = \phi_{\alpha\beta}\left(\bfx\right), \label{eq:ga1}
	\end{align}\label{def:ga}
	where $Id$ corresponds to the identity of the group $G$.
\end{mydef}

\begin{mydef}\label{def:inv}
	Let $\phi_{\cdot}(\cdot)$ be defined as in \eqref{eq:ga0}-\eqref{eq:ga1}. We say that   system \eqref{eq:dyn}-\eqref{eq:mes} is totally invariant under the action of $\phi_{\cdot}(\cdot)$  if
	\begin{enumerate}
		\item the dynamics are equivariant under $\phi_{\cdot}(\cdot)$, i.e.,
		\begin{align}
		\forall \alpha,\bfx,\bfu ,\bfw \quad \phi_{\alpha} \left(f(\bfx,\bfu,\bfw)\right) = f \left( \phi_{\alpha}(\bfx),\bfu ,\bfw \right), \label{eq:f_invariant}
		\end{align}
		\item the observation map $h(\cdot)$  is invariant w.r.t. $\phi_{\cdot}(\cdot)$ 
		\begin{align}
		\forall \alpha,\bfx \quad h(\phi_{\alpha}(\bfx), \bfv) = h(\bfx, \bfv). \label{eq:h_invariant}
		\end{align}
	\end{enumerate}	
\end{mydef}
``Symmetry" is defined as invariance to transformations $\phi_\alpha(\cdot)$.  Throughout the paper we will rather use the term  \emph{invariant}, along the lines of the preceding definition.

As in this paper we pursue the design of consistent EKFs, we will focus on the system ``seen" by an EKF: it consists of the linearization of system \eqref{eq:dyn}-\eqref{eq:mes} about the estimated trajectory $(\hat\bfx_n)_{n\geq 0}$ in the state space.  Along the lines of \cite{huang_analysis_2008, huang_observability-based_2010} we use the linearized system about a trajectory.

Let $\left(\bfx_n\right)_{n\geq 0}$ denote a solution of \eqref{eq:dyn} with noise turned off.  The local observability matrix  \cite{chen_1990_local} at $\bfx_{n_0}$ for the time interval between time-steps $n_0$  and $n_0+N$ is defined as
 \begin{align}
 \calO(\bfx_{n_0}) = \begin{bmatrix}
 \bfH_{n_0} \\
 \bfH_{n_0+1} \bfF_{n_0+1} \\
 \vdots \\
 \bfH_{n_0+N} \bfF_{n_0+N} \cdots \bfF_{n_0+1}
 \end{bmatrix}, \label{eq:ker}
 \end{align}
with the Jacobians $\bfF_n=\frac{\partial f}{\partial \bfx}|_{\bfx_{n-1},\bfu_{n},\bfw_{n}} $, $\bfH_n=\frac{\partial h}{\partial \bfx}|_{\bfx_{n}, \bfv_n}$.

First we show  the directions spanned by the action of $G$ are necessarily unobservable directions of the linearized system, that is, they lie in the kernel of the observability matrix.

\begin{mytheorem}\label{thm232}
	 If  system \eqref{eq:dyn}-\eqref{eq:mes} is invariant in the sense of Definition \ref{def:inv}, then the directions  $\frac{\partial}{\partial\alpha}\phi_{\alpha}|_{ Id}(\bfx_{n_0})$ infinitesimally  spanned by $\phi_{\cdot}(\cdot)$ at any  $\bfx_{n_0}$   necessarily lie in   $ \Ker
\calO(\bfx_{n_0})$, with  $\calO(\bfx_{n_0})$ defined by \eqref{eq:ker}, and are thus unobservable.

\begin{proof}
Differentiating\footnote{ On Lie groups differentiation can indeed be rigorously defined as $\frac{\partial}{\partial\alpha}\phi_{\alpha}\bigr |_{ \alpha=Id}(\bfx) \delta\alpha:=\frac{d}{ds}\phi_{\exp(s\delta\alpha)}(\bfx) |_{s=0}$ with $\delta\alpha$ in the Lie algebra, which mean partial derivative of $\phi_{ \alpha}\left(\bfx\right)$ with respect to $\alpha$ at $\alpha = Id$. \label{footnote}} \eqref{eq:f_invariant} and \eqref{eq:h_invariant} w.r.t. $\alpha$ at $Id$ we obtain
		\begin{align}
		\frac{\partial}{\partial\alpha}\phi_{\alpha}\bigr |_{ \alpha=Id}\left(f(\bfx,\bfu,\bfw)\right) &= \frac{\partial f}{\partial \bfx}|_{(\bfx,\bfu,\bfw)}\frac{\partial}{\partial\alpha}\phi_{\alpha}\bigr |_{ \alpha=Id}(\bfx), \label{eq:f_alpha}\\
		\frac{\partial h}{\partial \bfx}|_{(\bfx,\bfv)}\frac{\partial}{\partial\alpha}\phi_{\alpha}\bigr |_{ \alpha=Id}(\bfx) &= \bfzero\quad\forall \bfx\in\calX. \label{eq:h_alpha}
		\end{align}
		Let $\left(\bfx_n\right)_{n\geq 0}$ denote a solution of \eqref{eq:dyn} with noise turned off.   \eqref{eq:h_alpha} applied at $\bfx_{n_0}$ yields $\bfH_{n_0} \frac{\partial}{\partial\alpha}\phi_{\alpha}|_{Id}(\bfx_{n_0})    =\allowbreak \bfzero$. Considering then \eqref{eq:f_alpha} at $\bfx_{n_0}$ leads to $
		  \frac{\partial}{\partial\alpha}\phi_{\alpha}|_{ Id} \left(f(\bfx_{n_0},\bfu_{n_0+1},\bfw_{n_0+1})\right)  =\allowbreak\bfF_{n_0+1}\frac{\partial}{\partial\alpha}\phi_{\alpha}|_{ Id} \left(\bfx_{n_0}\right)$. 
		Applying   \eqref{eq:h_alpha} at $\bfx_{n_0+1} \allowbreak =f(\bfx_{n_0},\bfu_{n_0+1},\bfw_{n_0+1})$ yields  $
		\bfH_{n_0+1}\frac{\partial}{\partial\alpha}\phi_{\alpha}|_{ Id} \left(\bfx_{n_0+1}\right)\allowbreak  =\bfzero,
$ and thus $\bfH_{n_0+1}\allowbreak\bfF_{n_0+1}\frac{\partial}{\partial\alpha}\phi_{\alpha}|_{ Id}\left(\bfx_{n_0}\right)=\bfzero$. A simple recursion proves   $\frac{\partial}{\partial\alpha}\phi_{\alpha}|_{ Id} \left(\bfx_{n_0}\right)\subset\Ker \calO(\bfx_{n_0})$.	\end{proof}
\end{mytheorem}

Indeed, no matter the number of observations and moves, we are  inherently unable to detect an initial (infinitesimal) transformation $\phi_\alpha(\cdot)$, the problem being \emph{invariant} to it.

\subsection{Observability Issues of the Standard EKF}

Let $(\hat\bfx_n)_{n\geq 0}$ be a sequence of state estimates given by an EKF. The linearized system ``seen" by the EKF involves  the \emph{estimated} observability matrix
\begin{align}
\hat{\calO} (\hat \bfx_{n_0})= \begin{bmatrix}
\hat{\bfH}_{n_0} \\
\hat{\bfH}_{n_0+1} \hat{\bfF}_{n_0+1} \\
\vdots \\
\hat{\bfH}_{n_0+N} \hat{\bfF}_{n_0+N} \cdots \hat{\bfF}_{n_0+1}
\end{bmatrix}, \label{eq:ker2}
\end{align}where Jacobians are computed \emph{at the estimates}. The directions   spanned by $\phi_{\cdot}(\cdot)$ at   $\hat \bfx_{n_0}$  necessarily lie in   $ \Ker
\calO(\hat \bfx_{n_0})$, as proved by Proposition \ref{thm232}, \emph{but} there is a null probability  that they lie in $ \Ker\hat 
\calO(\hat \bfx_{n_0})$, because of the noise and  Kalman updates,  see \cite{huang_observability-based_2010}. A major consequence is that the EKF gains spurious information along the unobservable directions.

In fact,  this problem
 stems from the   choice of  estimation error $\bfx-\hat\bfx$ that does not match unobservability of system \eqref{eq:dyn}-\eqref{eq:mes}: changing the estimation error  may resolve the problem.

\subsection{EKF Based on a Nonlinear Error}\label{sec:new}
In this section, we define an EKF based on a  nonlinear function $\eta(\bfx,\hat\bfx)\in\calX$ that provides an alternative to the usual linear estimation error $\bfx-\hat\bfx$. We prove consistency under  compatibility assumptions of the group action and  $\eta(\cdot,\cdot)$. 

The methodology builds upon the alternative errors \begin{align}
\bfe_{n-1|n-1} &=  {\eta}\left(\bfx_{n-1},\hbfx_{n-1|n-1}\right) \label{eq:err0}, \\
\bfe_{n|n-1} &=  {\eta}\left(\bfx_{n}, \hbfx_{n|n-1}\right) \label{eq:err}.
\end{align}

The filter is displayed in Algorithm \ref{alg:ekf}. As  the covariance matrix $\bfP^e$ is supposed to reflect the dispersion of $\bfe$, we need to define Jacobians w.r.t our alternative state error.  At line 2 
 $\hat \bfF_n^e$, $\hat \bfG_n^e$ are Jacobians of the error propagation function,   and at line 3, $\hat \bfH_n^e$, $\hat \bfJ_n^e$ are Jacobians of the error measurement defined through the following first order approximations
\begin{align}
\bfe_{n|n-1} &\simeq \hat \bfF_n^e \bfe_{n-1|n-1} + \hat \bfG_n^e \bfw_n, \label{fodar}\\ 
\bfy_n - h\left(\hbfx_{n|n-1},\bfzero\right) &\simeq \hat \bfH_n^e \bfe_{n|n-1} + \hat \bfJ_n^e \bfv_n.\label{fodar2}
\end{align}

At line 4, $\bfe_n^+ $ denotes the  (best) error estimate according to the EKF. However, defining the (best) state corresponding  to estimate $\hbfx_{n|n}$  is not straightforward as in the linear case where $\hbfx_{n|n}=\hbfx_{n|n-1}+\bfe_n^+ $. At  line 5 we use a retraction $\psi: \calX\times \bbR^q \rightarrow \calX$, that is, \emph{any} function $\psi\left(\cdot \right)$ which is consistent with the error to the first  order, i.e.,  $\bfe_n^+\approx\eta(\hbfx_{n|n},\hbfx_{n|n-1})$.

Note that, we recover the conventional EKF   if we let  $\bfe=\eta\left(\bfx,\hbfx\right) = \bfx-\hbfx$ be the usual linear error.

\begin{algorithm}[h]
	\KwIn{initial estimate $\hbfx_0$ and uncertainty matrix $\bfP_0^e$}
	\While{filter is running}{
		\SetKwBlock{Propagation}{Propagation}{end}
		\Propagation{
			\nl $\hbfx_{n|n-1} = f\left(\hbfx_{n-1|n-1},\bfu_n,\bfzero\right)$\;
			\nl $\bfP^e_{n|n-1} = \hat \bfF_n^e \bfP_{n-1|n-1}^e  (\hat \bfF_n^e)^T + \hat \bfG_n^e \bfQ_n (\hat \bfG_n^e)^T$\;
		}
		\SetKwBlock{Update}{Update}{end}
		\Update{
			
			\nl $\bfK_n = \hat \bfH_n^e \bfP_{n|n-1}^e \allowbreak / \left(\hat \bfH_n^e \bfP_{n|n-1}^e (\hat \bfH_n^e)^T + \hat \bfJ_n^e \bfR_n  (\hat \bfJ_n^e)^T\right)$\;
			\nl $\bfe_n^+ = \bfK_n \left( \bfy_n - h\left(\hbfx_{n|n-1},\bfzero\right)\right)$\;
			\nl $\hbfx_{n|n} = \psi\left(\hbfx_{n|n-1},\bfe_n^+\right)
			$; \tcp{state update}
			\nl $\bfP_{n|n}^e = \left(\bfI-\bfK_n\hat{\bfH}_n^e\right) \bfP_{n|n-1}^e$\;
		}		
	}
	\caption{EKF based on a non-linear state error \label{alg:ekf}}
\end{algorithm}

\subsection{Compatibility Assumptions and Main Consistency Result}

The matrix\footnote{To fix ideas and help the reader understand the tools, let us pick an example. For instance we can let the state space be $\calX=\mathbb R^2$ and $\phi_\alpha(\cdot)$ a rotation of angle $\alpha$ around the origin. The group $G=S^1$ is the circle, and the identity element is $Id=0+2k\pi$. In this case, using polar coordinates, i.e. $\bfx=\left[r\cos\theta,r\sin\theta\right]^T$ we have $\phi_\alpha(r\cos\theta,r\sin\theta)=\left[r\cos(\theta+\alpha),r\sin(\theta+\alpha)\right]^T$. The directions spanned by $\phi_{\cdot}(\cdot)$ at $\bfx$ are $\frac{\partial}{\partial\alpha}|_{Id}\phi(\bfx)=\left[-r\sin\theta,r\cos\theta\right]^T$ which is a vector orthogonal to $\bfx$ in $\mathbb R^2$. Now, consider an error $\eta(\cdot,\cdot)$ between two elements of the state space $\calX$. What we advocate in the preset paper is that nonlinear errors may be much better suited for EKF design, but to keep things simple at this stage assume merely that $\eta(\cdot,\cdot)$ denotes the linear error. In this case we have $\eta(\phi_\alpha(\bfx),\bfx)=\phi_\alpha(\bfx)-\bfx$.  This is an element of the state space $\mathbb R^2$, and we can differentiate with respect to $\alpha$. We find $\frac{\partial}{\partial\alpha}|_{Id}\eta(\phi_\alpha(\bfx),\bfx)=\left[-r\sin\theta,r\cos\theta\right]^T$. Of course, it would have been different if we had chosen a different error. If we choose the polar coordinates error, that is $\eta(\cdot,\cdot)$ is a two component vector with first component the difference of norms and second component the difference of angles in polar coordinates, then we find $\eta(\phi_\alpha(\bfx),\bfx)=\left[0,\alpha\right]^T$ which is quite different, and thus $\frac{\partial}{\partial\alpha}|_{Id}\eta(\phi_\alpha(\bfx),\bfx)=\left[0,1\right]^T$. We see to some extent this error is more ``compatible'' with the symmetry group we chose since it is the same at any $\bfx\in\mathbb R^2$. In this footnote, the group was of dimension 1, i.e. it was encoded by a one dimensional element $\alpha$. If $\alpha$ had another component, we would need to differentiate with respect to it also. As a result we would not obtain a vector $\frac{\partial}{\partial\alpha}|_{Id}\eta(\phi_\alpha(\bfx),\bfx)$ but a matrix, each column being associated to a component of $\alpha$. }  $\frac{\partial}{\partial\alpha}|_{Id}\eta(\phi_\alpha(\bfx),\bfx)$ reflects how infinitesimal transformations of the state produced by the action of $G$ affect the error variable $\bfe=\eta(\cdot,\cdot)$. In Assumption \ref{ass1} below this matrix is used to define a kind of “compatibility” between an invariance group $G$ and a nonlinear error function $\eta(\cdot,\cdot)$, which leads to the main result of this paper (Theorem \ref{th:shift}).

\begin{ass}\label{ass1}
The image of the matrix $\frac{\partial}{\partial\alpha}|_{Id}\eta(\phi_\alpha(\bfx),\bfx)$ is a fixed  subspace $\calC$ that does not depend on $\bfx$. 
\end{ass}

 Proposition \ref{thm232} proved that if the system \eqref{eq:dyn}-\eqref{eq:mes} is totally invariant, then the directions  infinitesimally spanned by $\phi_{\cdot}(\cdot)$ at any point $\bfx_{n_0}$  lie in   $ \Ker
\calO(\bfx_{n_0})$, with  $\calO(\bfx_{n_0})$ defined by \eqref{eq:ker}, and thus they are unobservable. We have also recalled this  is not  true for the linearized system seen by the EKF, i.e. for $\hat \calO(\hbfx_{n_0})$   \cite{huang_observability-based_2010}.  We have the following powerful result:

\begin{mytheorem2}\label{th:shift}	If the system \eqref{eq:dyn}-\eqref{eq:mes} is invariant,  and under Assumption \ref{ass1}, the unobservable directions $\calC$ spanned by $\phi_{\cdot}(\cdot)$ at any point $\hat \bfx_{n_0}$,  \emph{measured using error $\eta(\cdot,\cdot)$} necessary    lie in   $ \Ker\hat \calO^e(\hat\bfx_{n_0})$, with $(\hat\bfF_n^e,\hat\bfH_n^e )_{n\geq n_0}$ defined by \eqref{fodar}-\eqref{fodar2}.

\begin{proof}
	Let $\hat \bfM_{n}:= \frac{\partial}{\partial\alpha}|_{Id}\eta(\phi_\alpha(\hat\bfx_{n}),\hat\bfx_{n})$.	Recalling \eqref{fodar2}, we have   $h(\phi_{\alpha}(\hat\bfx_{n}))-h(\hat\bfx_{n})\simeq\hat \bfH_n^e\eta(\phi_{\alpha}(\hat\bfx_{n}),\hat\bfx_{n})\simeq\hat\bfH_n^e\hat\bfM_n\delta\alpha$ with $\delta\alpha$ a linearized approximation to $\alpha\in G$.  But also $h(\phi_{\alpha}(\hat\bfx_{n}))-h(\hat\bfx_{n})\simeq\frac{\partial h}{\partial \bfx}|_{(\hat \bfx_n,\bfzero)} \frac{\partial}{\partial\alpha}\phi|_{Id}(\hat\bfx_{n}) \delta\alpha$. From  \eqref{eq:h_alpha} the latter is $\bfzero$.  Thus $\hat \bfH_n^eu=\bfzero$ for any $u\in\calC$, i.e. $\hat \bfH_n^e\calC=\bfzero$.

Using \eqref{fodar}, $\hat \bfF^e$	 is defined as	
\begin{align}
\eta\left(f(  \bfx), f(\hat\bfx)\right)  \simeq \hat \bfF^e \eta\left(\bfx  , \hat\bfx \right), 
\end{align}
thus $\eta(f(\phi_{\alpha}(\hat\bfx_{n})),f(\hat\bfx_{n}))\simeq \hat \bfF_{n+1}^e \eta(\phi_{\alpha}(\hat\bfx_{n}),\hat\bfx_{n})$ $\simeq\hat \bfF_{n+1}^e   \hat\bfM_n\delta\alpha$. 
Besides, using \eqref{eq:f_invariant} yields 
\begin{align}
\eta(f(\phi_{\alpha}(\hat\bfx_{n})),f(\hat\bfx_{n})) &=\eta(\phi_{\alpha}(f(\hat\bfx_{n})),f(\hat\bfx_{n})) 
\\&\simeq  \frac{\partial}{\partial\alpha}|_{Id}\eta(\phi_\alpha(f(\hat\bfx_{n})),f(\hat\bfx_{n}))\delta\alpha\in\calC
\end{align}
applying Assumption \ref{ass1} at $f(\hat\bfx_{n})$. Thus  for any $\delta\alpha$ we have $\hat \bfF_{n+1}^e\hat\bfM_n\delta\alpha\subset\calC$ and thus  $\hat \bfF_{n+1}^e\calC\subset \calC$. We have thus  proved\begin{align}\hat \bfH_n^e\calC=\bfzero,\quad \hat \bfF_{n+1}^e\calC\subset \calC,\quad\text{ for any $n$ and $\hat\bfx_n$.
}\label{tulesai}
\end{align}This proves the result through an immediate recursion.
	\end{proof}
\end{mytheorem2}

We obtain the consistency property we pursue:
the linearized model has the desirable property of the linear Kalman filter regarding the unobservabilities, when expressed in terms of error $\eta(\cdot,\cdot)$. As a byproduct, the
unobservable subspace seen by the filter is automatically of
appropriate dimension.

\subsection{Consequences in Terms of  Information}
In the linear
Gaussian case, the inverse of the covariance matrix output by
the Kalman filter is the Fisher information available to the
filter (as stated in \cite{bar-shalom_state_2002} p. 304). Thus, the
inverse of the covariance matrix $\bfP_{n|n}^{-1}$ output by any EKF should 
reflect an absence
of information gain along unobservable directions. Otherwise, the output covariance matrix would be too optimistic, i.e., the filter inconsistent \cite{bar-shalom_state_2002}. 
 
Note that, the covariance matrix $\bfP_{n|n}^e$ \emph{reflects the dispersion of the error}  $\bfe_{n|n}$ of \eqref{eq:err0}-\eqref{eq:err}, as emphasized by the superscript $e$.  We have the following consistency result.

\begin{mytheorem}\label{mainthm}
	Let $u_{n_0}\in Im~ \frac{\partial}{\partial\alpha}|_{Id}\eta(\phi_\alpha(\hat\bfx_{n_0}),\hat\bfx_{n_0})$ be an unobservable direction  spanned by $\phi_{\cdot}(\cdot)$ at the estimate $\hat\bfx_{n_0}$, measured using alternative error $\eta(\cdot,\cdot)$. Let  $(u_n)_{n\geq n_0}$ with $u_{n }=\hat\bfF_{n }^e u_{n-1}$ be  its propagation through the linearized model. Under  Assumption \ref{ass1} the Fisher information according to the filter about $(u_n)_{n\geq n_0}$   is non-increasing , i.e., 	\begin{align}
	   u_{n }^T (\bfP_{n|n}^e)^{-1}  u_{n} \leq    u_{n-1}^T (\bfP_{n-1|n-1}^e)^{-1}   u_{n-1}.	
	\end{align}

	\begin{proof}
At propagation step we have 
\begin{align}
u_{n }^T (\bfP^e_{n |n-1})^{-1} u_{n} &= u_{n-1}^T(\hat\bfF_n^e)^T ( \hat\bfF_n^e \bfP^e_{n-1|n-1}(\hat\bfF_n^e)^T+\hat\bfG_n^e \bfQ_n (\hat\bfG_n^e)^T )^{-1}\hat\bfF_n^e u_{n-1} \\
&\leqslant u_{n-1}^T (\hat\bfF_n^e)^T \left( \hat\bfF_n^e \bfP^e_{n-1|n-1}(\hat\bfF_n^e)^T \right)^{-1}\hat\bfF_n^e u_{n-1}
\end{align}
  since $\bfQ$ is positive semidefinite. As $(\hat\bfF_n^e)^{-1}\hat\bfF_n^e=\bfI$ we have just proved $u_{n}^T (\bfP^e_{n|n-1})^{-1} u_{n}\leqslant u_{n-1}^T (\bfP^e_{n-1|n-1})^{-1} u_{n-1}^T$. 
 
 At   update  step (in information form) we have
 \begin{align}
 u_{n}^T (\bfP^e_{n|n})^{-1} u_{n } = u_{n }^T \left( (\bfP^e_{n |n-1})^{-1} + (\hat\bfH_{n }^e)^T \hat\bfR_n^{-1} \hat\bfH_{n }^e \right) u_{n }.\end{align} But using \eqref{tulesai} we see  $u_i\in\calC$ $\forall i\geq n_0$ and thus $\bfH_{n }^eu_{n }=\bfzero$ so $u_{n}^T (\bfP^e_{n|n})^{-1} u_{n} = u_{n }^T (\bfP^e_{n|n-1})^{-1} u_{n } \leqslant u_{n-1}^T(\bfP^e_{n-1|n-1})^{-1}u_{n-1}$.
%
%
%
\end{proof}
\end{mytheorem}
The theorem essentially ensures the \emph{linearized model} of the
filter has a structure which guarantees that the covariance
matrix at all times reflects an absence of ``spurious" (Bayesian
Fisher) information gain over unobservable directions, ensuring strong consistency properties of our alternative EKF.

\section{Application to Multi-Sensor Fusion for  Navigation}\label{nav:ss}
In this section, we consider a navigating vehicle or a robot  equipped with sensors which only measure quantities relative to the vehicle's frame. Thus the vehicle cannot acquire information about its absolute position and orientation, which results in inevitable unobservability. 
The   state   space is $\calX= SO(3) \times \bbR^{3l+3m+k}$ and the state $\bfx$ is defined as
\begin{align}
	\bfx = \left(\bfR,~ \bfp_1,~ \cdots,~ \bfp_l,~ \bfv_1,~ \cdots,~ \bfv_m,~ \bfb\right) \in \calX, \label{eq:defx}
\end{align}
where $\bfR \in SO(3)$ represents the orientation of the vehicle, i.e., its  columns are the axes of the  vehicle's frame, and where
\begin{enumerate}
	\item $\lbrace\bfp_i \in \bbR^3\rbrace_{i=1,\ldots,l}$ are vectors of the global frame, such as the vehicle's position,
	\item $\lbrace\bfv_i \in \bbR^3\rbrace_{i=1,\ldots,m}$ are velocities in the global frame, and higher order derivatives of the $\bfp_i$'s.
	\item  $\lbrace\bfb\rbrace\in \bbR^k$, are quantities being  invariant to global transformations, such as sensors' biases or camera's calibration  parameters. 
\end{enumerate}
Without restriction, we consider in the following $l=m=1$ for  convenience of notation.

\begin{mydef}
The Special Euclidean group $SE(3)$ describes   rigid motions in 3D and is defined as $SE(3) = \lbrace \alpha=(\bfR_\alpha, \bfp_\alpha), \bfR_\alpha \in SO(3), \bfp_\alpha \in \bbR^3\rbrace$. Given $\alpha, \beta \in SE(3)$, the group operation is $\alpha \beta = (\bfR_\alpha \bfR_\beta, \bfR_\alpha\bfp_\beta + \bfp_\alpha)$ and the inverse $\alpha^{-1} = (\bfR^T_\alpha, -\bfR_\alpha^T\bfp_\alpha)$. We denote  $Id$ the identity.  
\end{mydef}
Changes of global frame are encoded as the action $\phi_\alpha(\bfx)$ of an element $\alpha \in SE(3)$ on $\calX$. Thus, quantities expressed in the global frame (such as the vehicle position) are rotated and translated by the action $\phi_\alpha(\cdot)$, whereas quantities expressed in the vehicle's frame, such as Inertial Measurement Unit (IMU) biases,  are left unchanged. The action then writes  (with  $ i=1,\ldots, l$ and $j=1,\ldots,  m$)\begin{align}
\begin{split}
\phi_\alpha(\bfx) = \big(&\bfR_\alpha \bfR, ~\bfR_\alpha \bfp_i + \bfp_\alpha,~ \bfR_\alpha \bfv_j, ~ \bfb \big).\label{eq:defphi}
\end{split}
\end{align}

\begin{ass}\label{ass2} The vehicle's dynamic does not depend on the choice of global frame, and  the vehicle's sensors  only have access to relative observations, i.e. no absolute information  is available. 
\end{ass}

As a result, the  equations write  \eqref{eq:dyn}-\eqref{eq:mes} and are invariant to the action of $SE(3)$ in the sense of Definition  \ref{def:inv}.

To differentiate w.r.t. elements of $ SE(3)$ we resort to its Lie algebra and do in detail what is  sketched in Footnote \ref{footnote}.

\begin{mydef}
The Lie algebra $\mathfrak{se}(3)$ of $SE(3)$ encodes small rigid motions about the identity. It is defined as $\lbrace\left((\delta\boomega)_\times, \delta \bfp\right);~\delta\boomega,\delta \bfp\in\mathbb R^3\rbrace$, where $(\boomega)_\times$ is the skew symmetric matrix associated with cross product with $\boomega\in\mathbb R^3$. For any $\delta\alpha\in\mathfrak{se}(3)$, we have $\alpha:=\exp_{SE(3)}(\delta\alpha)\in SE(3)$ where $\exp_{SE(3)}(\cdot)$ denotes the exponential map of $SE(3)$ (for a   definition see \eqref{exp3}-\eqref{exp1} below  with $l=1,m=0,k=0$).
\end{mydef}For more information about   $SE(3)$ and its use in state estimation for robotics see the recent monographs  \cite{barfoot_state_2017,chirikjian2011stochastic}.

Writing $(\bfR_\alpha,\bfp_\alpha)=\exp_{SE(3)}\left(\left[(\delta\boomega)_\times, \delta \bfp\right]\right)$ in \eqref{action:slam}  we see the directions infinitesimally spanned by $\phi_{\cdot}(\cdot)$ at   $\bfx$ in the direction  $\delta\alpha=\left((\delta\boomega)_\times, \delta \bfp\right)$   
	write for the  state \eqref{eq:defx}:
	\begin{align}
	\begin{split}
 \frac{\partial}{\partial \alpha}\phi_{\alpha}|_{Id}(\bfx) \delta \alpha= \big(& (\delta \boomega)_\times \bfR, ~
	(\delta \boomega)_\times \bfp_i + \delta \bfp, ~
	(\delta \boomega)_\times \bfv_j,~	\bfb \big)
	\end{split}\label{unobsdirr}
	\end{align}
	where $ i=1,\ldots, l$ and $j=1,\ldots,  m$.

\begin{example}\label{ex:slam}[SLAM]
	Consider a simple SLAM system with one robot and one landmark \cite{huang_observability-based_2010}. Let $\bfp_R$ be the position of the robot, $\bfR$   the orientation of the robot, and $\bfp_L$ the landmark's position. The state  is  
	\begin{align}
	\bfx = \left(
	\bfR,~ \bfp_R,~ \bfp_L
	\right) \in \calX = SO(3) \times \bbR^6.
	\end{align}
	The dynamics write $f(\bfx,\bfu,\bfw)=(\bfR\bar\bfR,  \bfp_R +\bfR \bar\bfR \bar \bfp, \bfp_L)$ where $\bar\bfR,\bar \bfp$ denote orientation and position increments typically measured through odometry. The observation of the landmark in the robot's frame is of the form $\bfy = \tilde h(\bfR^T(\bfp_L-\bfp_R))$.  Translations and rotations of the global frame correspond to actions of  elements   $\alpha = (\bfR_\alpha, \bfp_\alpha)\in SE(3)$ as  
	\begin{align}
	\phi_\alpha(\bfx) = \left(
	\bfR_\alpha \bfR,~ \bfR_\alpha \bfp_R + \bfp_\alpha,~ \bfR_\alpha \bfp_L + \bfp_\alpha\right).\label{action:slam}
	\end{align}
	The system is obviously invariant. 
	
	Referring to \eqref{unobsdirr}, as $l=2,m=0$ and $\bfp_R=\bfp_1,\bfp_L=\bfp_2$,   the directions spanned by $\phi_{\cdot}(\cdot)$ at   $\bfx$  are as follows
	\begin{align}
	\left(
	\left(\delta \boomega\right)_\times \bfR, ~(\delta \boomega)_\times \bfp_R + \delta \bfp,~ (\delta \boomega)_\times \bfp_L + \delta \bfp
	\right),\label{direct;}
	\end{align}with $\left((\delta\boomega)_\times, \delta \bfp\right)\in\mathfrak{se}(3)$. As a direct consequence of  Prop. \ref{thm232}, those directions are  unobservable. The system continues to be invariant even if a  sophisticated model is assumed: the motion equations do not depend a choice of  global frame. 
\end{example}

\begin{example}\label{ex:slam23}[VIO, or Visual Inertial Navigation System (VINS)]\label{ex:vio}
	Consider a  vehicle equipped with an IMU and a camera, as in \cite{li_high-precision_2013}. Let $\bfv$ denote the vehicle velocity, $\bfb$ the IMU bias and/or scale factors. The state  is
	\begin{align}
	\bfx = \left(\bfR,~ \bfp_R,~  \bfv,~ \bfb\right)
	\end{align}
	and observations correspond to landmarks' bearings in the vehicle frame, whereas   inputs $\bfu_n \in \bbR^6$ are provided by an IMU.
	A change of global frame  $\alpha = (\bfR_\alpha, \bfp_\alpha)$ writes 	\begin{align}
	\phi_\alpha(\bfx) = \left(
	\bfR_\alpha \bfR,~ \bfR_\alpha \bfp_R + \bfp_\alpha,~  \bfR_\alpha \bfv,~ \bfb
	\right),
	\end{align}
	where we restrict $\bfR_\alpha$ to be around the gravity axis, i.e. with $\bfR_\alpha \bfg =  \bfg $, since the vertical is measured \cite{li_high-precision_2013}. The action $\phi_{\cdot}(\cdot)$ is then also invariant.
	
\end{example}

\subsection{EKF Based on a Nonlinear Error}
 For the general state $\bfx$ of \eqref{eq:defx}  consider the  nonlinear  error\begin{align}
\begin{split}
\eta(\bfx,\hbfx) = \big(&\bfR \hbfR^T, \hbfp_i- \hbfR\bfR^T \bfp_i
,\hbfv_j- \hbfR\bfR^T \bfv_j
,\bfb-\hbfb \big)\label{eq:eta}
\end{split}.
\end{align}where $ i=1,\ldots, l$ and $j=1,\ldots, m$. We set $l=m=1$ for simplicity. 
To linearize, we have the following  first order vector approximation  of   \eqref{eq:eta}  that lives in $ \bbR^{3(1+l+m)+k}$
\begin{align}
\begin{split}
\check{\eta}(\bfx, \hbfx) &= \big(\bfe_R
,~\hbfp- (\bfe_R)_\times \bfp
,~\hbfv- (\bfe_R)_\times \bfv 
,~ \bfb-\hbfb \big),  
\end{split} \\
\bfR \hbfR^T &= \exp_{SO(3)}(\bfe_R )\simeq \bfI + (\bfe_R)_\times + o(\|\bfe_R\|^2),
\end{align}
{ where $\bfe_R \in \bbR^3$.}
\begin{mytheorem}\label{th:inv}
	The error  \eqref{eq:eta}  is compatible with the action \eqref{eq:defphi} of $SE(3)$ in the sense of Assumption \ref{ass1}.

	\begin{proof}
		For $\alpha \in SE(3)$, using \eqref{eq:eta} we obtain that 
		\begin{align}
		\eta(\phi_{ \alpha}(\bfx),\bfx) &= \big(\bfR_\alpha \bfR \bfR^T
		,~\bfp  - \bfR (\bfR_\alpha \bfR)^T  [\bfR_\alpha \bfp+ \bfp_\alpha], \ldots \nonumber \\
		&~\bfv- \bfR (\bfR_\alpha \bfR)^T \bfR_\alpha \bfv
	   ,~ \bfb-\bfb
		\big)  	\\
		&=\big(\bfR_\alpha
		, -\bfR_\alpha^T\bfp_\alpha
		,~\bfzero_3
		,~\bfzero_k\big),		\end{align}
		 such that $\eta(\phi_{ \alpha}(\bfx),\bfx)$ turns out to be  \emph{independent} of $\bfx$, and the result is readily obtained by differentiation at $\alpha=Id$. 
	\end{proof}
\end{mytheorem}

\subsection{Choice of the Retraction}
When the state space is a Lie group, and one uses errors that are invariant with respect to right multiplication, the theory of Invariant EKF  (IEKF) \cite{bonnable2009invariant,barrau_invariant_2017} suggests to use $\psi\left(\bfx, \bfe\right) =\exp(\bfe)\bfx$ where $\exp(\cdot)$ denotes the Lie group exponential.  In \cite{bonnabel_symmetries_2012}, it was noticed a natural Lie group structure underlies the (odometry based) SLAM problem.  In the preprint  \cite{barrau_ekf-slam_2015} of the current article, we formalized more elegantly this group, and called it $SE_{l+1}(3)$. The current paper is a generalization to more complete state \eqref{eq:defx}, which may be endowed with the group structure $SE_{l+m}(3) \times \bbR^k$ (direct product of groups $SE_{l+m}(3)$ and $\bbR^k$). For state \eqref{eq:defx} we thus  suggest  $\bfx^+ = \psi\left(\bfx, \bfe\right) :=\exp_{SE_{l+m}(3) \times \bbR^k}(\bfe)\bfx$, i.e. \begin{align}
\begin{split}
\psi\left(\bfx, \bfe\right) =& \big(\delta \bfR^+\bfR,  
\delta \bfR^+\bfp_i + \delta \bfp_i^+, 
\delta \bfR^+\bfv_j+ \delta \bfv_j^+, 
\bfb + \delta\bfb^+\big)\end{split}
\end{align}
with  $ i =1,\ldots, l,~j =1,\ldots, m$, and where
\begin{align}
&\left[\begin{array}{c|cccccc}\arraycolsep=4pt
 \delta \bfR^+ & \delta \bfp_1^+& \cdots & \delta \bfp_l^+ & \delta \bfv_1^+ & \cdots & \delta \bfv_m^+ \\
 \hline
\bfzero &\multicolumn{6}{c}{\bfI_{3l+3m}}\end{array}\right] \label{exp3}  \\
& :=\bfI + \bfS + \frac{1-\cos(\|\bfe_R\|)}{\|\bfe_R\|^2}\bfS^2 + \frac{\|\bfe_R\|-\sin(\|\bfe_R\|)}{\|\bfe_R\|^3}\bfS^3, \label{exp2}
\end{align}
\begin{align} 
 \bfS &:= 	\left[
\begin{array}{c|cccccc} \arraycolsep=1.4pt
(\bfe_R)_\times & \bfe_{\bfp_1} & \cdots & \bfe_{\bfp_l}& \bfe_{\bfv_1}& \cdots & \bfe_{\bfv_m} \\
\hline
\bfzero &\multicolumn{6}{c}{\bfzero_{3l+3m}} 
\end{array}\right],\label{exp1}
\end{align}
$\delta \bfb^+ = \bfe_\bfb$, $\bfzero =\bfzero_{3+3m+3l \times 3}$ and $\bfI = \bfI_{3+3m+3l \times 3}$.

\subsection{Extension to Problems Involving Multiple Robots}\label{sec:extmul}
Consider a problem consisting of $M$ systems, see e.g. Section \ref{sec:exp}. We have $M$ global orientations, one for each system, that transform as the global frame's orientation. For such problems, we define a collection of $\bfx_i$ of  \eqref{eq:defx}, and the alternative state error of the problem $\eta(\cdot,\cdot)$ merely writes
\begin{align}
\eta\left(\bfx_1,\ldots,\bfx_M\right) = \left(\eta_1\left(\bfx_1,\hbfx_1\right),\ldots,\eta_M\left(\bfx_M,\hbfx_M\right)\right),
\end{align}
where $\eta_i\left(\bfx_i,\hbfx_i\right)$ is the error \eqref{eq:eta} for the $i$-th system.
\section{Simulation Results} \label{sec:sim}
\begin{figure}
	\centering
		\begin{tikzpicture}[]
	\begin{axis}[height=8.5cm, width=8.5cm, 
	xlabel={$x$ (m)},
	ylabel={$y$ (m)},
	legend entries={true trajectory, features,  sensor range},
	legend pos= outer north east,
	grid=both, ticks=both,xlabel style={xshift=-0ex,yshift=0.0cm}]
	\addplot+[ultra thick,smooth,mark=,draw = red] table[x=x,y=y] {fig_traj.txt};
	\addplot+[only marks, mark=o,color=blue] table[x=x_fea,y=y_fea] {fig_traj.txt};
	\addplot+[ultra thick,smooth,mark=+,draw = black, only marks] table[x=x_init,y=y_init] {fig_traj.txt};
	\addplot+[thick,smooth,mark=,draw = black] table[x=x_range,y=y_range] {fig_traj.txt};
	\end{axis}
	\end{tikzpicture}
	\caption{Simulated trajectory : the displayed loop is driven by	a robot able to measure the relative position of the landmarks lying in a range of \SI{5}{m} around it. Velocity is constant as well as angular velocity.\label{fig:traj}} 
\end{figure}
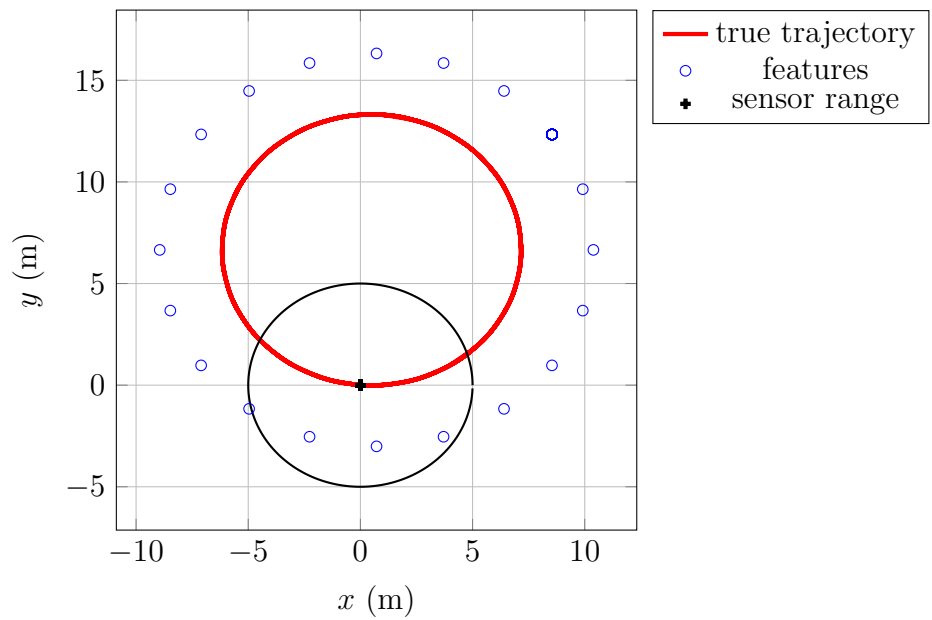

This section considers the 2D wheeled-robot SLAM problem and illustrates the performances of the proposed approach. We conduct similar numerical experiment as in the sound work \cite{huang_observability-based_2010} dedicated to EKF inconsistency  and   benefits of the OC-EKF, i.e., a robot makes 7 circular loops and 20 landmarks are disposed around the trajectory, see Figure \ref{fig:traj}. We refer the interested reader to the available Matlab code for the parameter setting and reproducing the present results.

\begin{figure}
	\centering
	{\footnotesize
		\begin{tabular}{c||c|c|c}
			\toprule
			Error &  EKF,\cite{huang_observability-based_2011} & \textbf{proposed},  see \eqref{eq:eta} & Robocent.\cite{castellanos_robocentric_2007}  \\
			\midrule
			orientation &  {$\hat\theta-\theta$} &  {$\hat\theta-\theta$} &{ $\hat\theta-\theta$} \\
			position & {$\hbfp_R-\bfp_R$} &  {$ \hbfp_R- \bfR(\hat\theta)\bfR(\theta)^T \bfp_R$} & {$ \bfR(\hat\theta)^T\hbfp_R- \bfR(\theta)^T \bfp_R$} \\
			landmark  & {$\hbfp_L-\bfp_L$} &  {$\hbfp_L- \bfR(\hat\theta)\bfR(\theta)^T \bfp_L$} & {$\bfR(\hat\theta)^T\left(\hbfp_L-\hbfp_R\right)\dots  $}\\\
			& &  & {\small$-\bfR(\theta)^T\left(\bfp_L-\bfp_R\right)$} \\
			\bottomrule
		\end{tabular}
	}
	\caption{Alternative state error definitions on the 2D SLAM problem for different solutions (extension of the approach  to 3D is immediate), with state $\bfx=(\theta,\bfp_R,\bfp_L)$ with $\bfp_R$ the position of the robot, $\mathbf{\theta}$ its orientation,  and $\bfp_L$ a landmark position.  $\bfR(\theta)$ denotes the planar rotation of angle $\theta$.\label{fig:error_def}} 
	
\end{figure}

We compare our approach to standard EKF which conveys an estimate of the linear error; 
 OC-EKF \cite{huang_observability-based_2010} which linearizes the model in a nontrivial  way to enforce the unobservable subspace of $\hat\calO$ to have an appropriate dimension;   robocentric EKF \cite{castellanos_robocentric_2007,guerreiro_globally_2013,lourenco_simultaneous_2016} which express \emph{the state   in the robot's frame} and then devise an EKF; and iSAM \cite{kaess_isam2:_2012,kummerle_g2o:_2011}, a popular  optimization technique  both for SLAM and odometry estimation which finds the most likely state trajectory given all past measurements. The differences between the   estimation errors used by the various EKF variants are recapped in Figure \ref{fig:error_def}. 

The results confirm the consistency guarantees of Theorem \ref{th:shift} and Proposition \ref{mainthm} are beneficial to the EKF in practice.

\subsection{Monte-Carlo Based Numerical Results}

Figure \ref{fig:cons} displays the Normalized Estimation Error Squared (NEES), Root Mean Square Error (RMSE) and distance to Maximum-Likelihood estimate, over 1000 Monte-Carlo runs.

\subsubsection{Consistency Evaluation}
\begin{figure}
	\centering
		\begin{tikzpicture}[]
	\begin{axis}[height=4.5cm, width=10.5cm, 
	ylabel={NEES},
	enlarge x limits=false,
	xmin = 0, xmax = 300,ymin = 0, ymax=3.1,
	ylabel style={xshift=0ex,yshift=-0.0cm},
	grid=both, ticks=both,xlabel style={xshift=-0ex,yshift=0.0cm}]
	\addplot+[thick,smooth,mark=,draw = yellow] table[x=t,y=isam] {fig2.txt};
	\addplot+[thick,smooth,mark=,draw = blue] table[x=t,y=ekf] {fig2.txt};
	\addplot+[ultra thick,smooth,mark=,draw = green] table[x=t,y=iekf] {fig2.txt};
	\addplot+[thick,smooth,mark=,draw = red] table[x=t,y=ocekf] {fig2.txt};
	\addplot+[thick,smooth,mark=,draw = cyan] table[x=t,y=rc] {fig2.txt};
	\addplot[thick, gray, densely dotted,forget plot] coordinates {(38,0)(38,3.5)};

	\end{axis}
	
	\begin{axis}[height=4.5cm, 
	width=10.5cm,
	xlabel style={xshift=-0ex,yshift=0.0cm},
	ylabel style={xshift=0ex,yshift=-0.0cm},
	at = {(0,-400)}, 
	ylabel={RMSE (m)},
	enlarge x limits=false,
	xmin = 0, xmax = 300,ymin = 0,ymax=3.3,
	label style={font=\small},
	grid=both,
	ytick= {0,1,2,3}]
	\addplot+[thick,smooth,mark=,draw = yellow] table[x=t,y=isam] {fig3_position.txt};
	\addplot+[thick,smooth,mark=,draw = blue] table[x=t,y=ekf] {fig3_position.txt};
	\addplot+[ultra thick,smooth,mark=,draw = green] table[x=t,y=iekf] {fig3_position.txt};
	\addplot+[thick,smooth,mark=,draw = red] table[x=t,y=ocekf] {fig3_position.txt};
	\addplot+[thick,smooth,mark=,draw = cyan] table[x=t,y=rc] {fig3_position.txt};
	\addplot+[smooth,dashed,mark=,draw = blue] table[x=t,y=ekf] {fig_3sigma.txt};
	\addplot+[smooth,dashed,mark=,draw = green] table[x=t,y=iekf] {fig_3sigma.txt};
	\addplot+[smooth,dashed,mark=,draw = red] table[x=t,y=ocekf] {fig_3sigma.txt};
	\addplot+[smooth,dashed,mark=,draw = cyan] table[x=t,y=rc] {fig_3sigma.txt};
	\addplot[thick, gray, densely dotted,forget plot] coordinates {(38,0)(38,3.5)};	
	\node[color=gray] at (90,2.7) {\scriptsize{$\leftarrow$ \texttt{first loop closure}}};
	\end{axis}
	
	\begin{axis}[height=4.5cm, width=10.5cm, xlabel={time steps},ylabel={distance to iSAM (m)},
	at = {(0,-290)},
	enlarge x limits=false,
	xlabel style={xshift=-0ex,yshift=0.0cm},
	ylabel style={xshift=0ex,yshift=-0.0cm},
	xmin = 0, xmax = 300, ymin = 0, ymax = 1.2,
	legend entries={iSAM \cite{kaess_isam2:_2012}, standard EKF,\textbf{proposed EKF},OC-EKF \cite{huang_observability-based_2010}, robocentric \cite{castellanos_robocentric_2007}},
	legend style={font=\scriptsize}, 
	legend columns=2,
	legend pos= north east,
	grid=both, ticks=both]
	\addplot+[smooth,thick,mark=,draw = yellow] table[x=t,y=ekf] {fig4.txt};
	\addplot+[smooth,thick,mark=,draw = blue] table[x=t,y=ekf] {fig4.txt};
	\addplot+[smooth,ultra thick,mark=,draw = green] table[x=t,y=iekf] {fig4.txt};
	\addplot+[smooth,thick,mark=,draw = red] table[x=t,y=ocekf] {fig4.txt};
	\addplot+[smooth,thick,mark=,draw = cyan] table[x=t,y=rc] {fig4.txt};	
	\addplot[thick, gray, densely dotted,forget plot] coordinates {(38,0)(38,1.5)};
	\end{axis}
	\end{tikzpicture}

	
	\caption{Average  performances of the different methods over 1000 runs. NEES  (for the robot 3-DoF pose) is the consistency indicator, and full consistency corresponds to NEES equal to 1. We see proposed EKF and iSAM are the more consistent, followed by OC-EKF and robocentric EKF, whereas standard EKF is not consistent.
		The accuracy is evaluated  in terms of the robot position RMSE. Standard EKF shows degraded performances as compared to others, which
		all achieve comparable performances. Finally, filters are evaluated in terms of
		average proximity of robot's estimated position with iSAM's, which computes the most likely state $\bfx_n$ given all past measurements $\bfy_1,\ldots,\bfy_n$. It is used as
		a reference of the best achievable estimate. We see that the proposed EKF
		 are the closest to iSAM. Dashed lines correspond to $3\sigma$ confidence upper bounds, and we see EKF is over-optimistic. Figures best seen in color.\label{fig:cons}}
\end{figure}
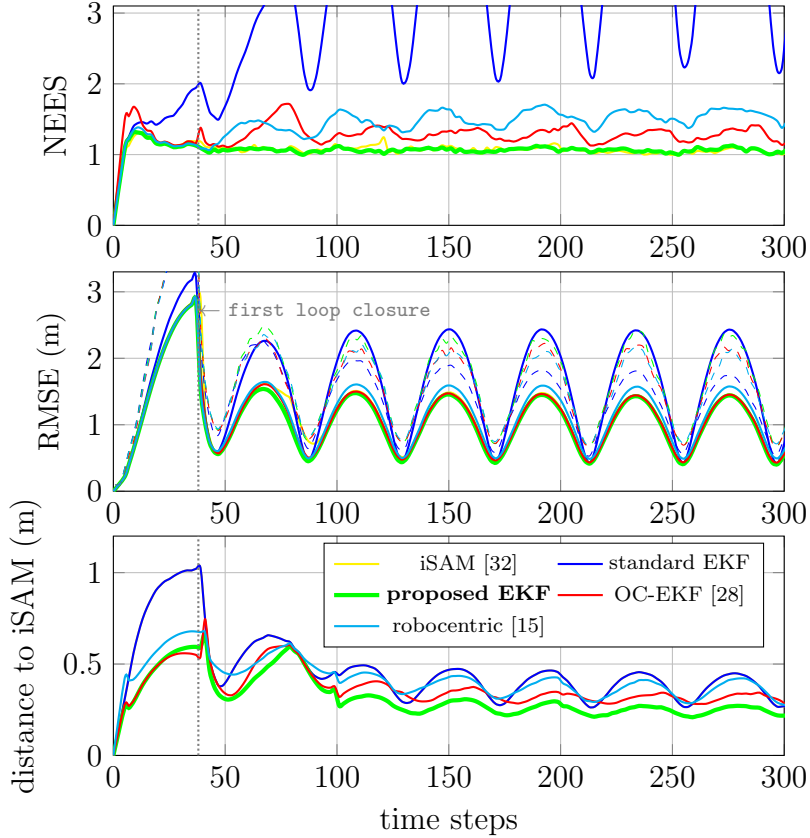

The NEES \cite{bar-shalom_state_2002} provides information about the filter  consistency, such that NEES $>$ 1 reveals an inconsistency issue: the actual uncertainty is higher than the computed uncertainty. As expected, the NEES of the robot pose  estimates in Figure \ref{fig:cons} indicates that the standard EKF is inconsistent,  whereas the other approaches are more consistent. The proposed EKF and iSAM obtain the best NEES, whereas the NEES of OC-EKF and robocentric EKF slightly increase after the first turn, i.e. at the first loop closure.
\subsubsection{Accuracy Performances}
We evaluate accuracy through RMSE of the robot position error. This confirms that: ``solving consistency issues improves the accuracy of the estimate as a byproduct, as wrong covariances yield wrong gains" \cite{bar-shalom_state_2002}. Numerical results are displayed in Figure \ref{fig:time_simu}.
\subsubsection{Distance to Maximum-Likelihood Estimate} We use
as a third performance criterion distance to the estimates
returned by iSAM \cite{kaess_isam2:_2012}, which are optimal in the sense that it returns
the Maximum A Posteriori (MAP) estimate. To this respect, we see that the {proposed EKF is the closest to iSAM.}  
\subsubsection{Execution Time} We provide the  execution time of the filters for the 100 Monte-Carlo runs in Table \ref{fig:time_simu}, which are implemented in Matlab and tested on Precision Tower 7910
armed with CPU E5-2630 v4 2.20 Hz. The iSAM's execution time is not included since it cannot be compared: it is implemented using C++ and an optimized code, whereas we used Matlab based simulations. It is thus evidently lower. Regarding computational complexity, our proposed filter has similar complexity as the standard EKF and OC-EKF, since its EKF-based structure makes it  quadratic in  the state dimension, i.e.,  number of landmarks. The use of a retraction at the update step instead of mere addition may slightly increase the computational burden, but the impact in the execution time proves negligible. The robocentric filter is penalized because it moves the landmarks during propagation, which in turn impacts the propagation of the covariance matrix. In our solution landmarks remain fixed during propagation. Note that the proposed solution can be implemented using recent techniques \cite{leung_decentralized_2012,huang_sparse_2008} to decrease computational load. To implement the  robocentric and OC-EKF we report that we used the  code of \cite{huang_observability-based_2010}, see Acknowledgments.  

These simulations confirm  that regarding SLAM, the proposed filter is an alternative to the OC-EKF. Contrarily to OC-EKF the model is linearized at the (best) estimate, and is thus much closer to standard EKF methodology, and applies to a large  class of problems without explicit computation of the unobservable directions.

\renewcommand{\figurename}{Table}
\setcounter{figure}{0} 
\begin{figure}
	\centering
	\begin{tabular}{c||c|c|c|c}
		\toprule
		Filter & EKF & \cite{huang_observability-based_2011} & \cite{castellanos_robocentric_2007} & \textbf{proposed}   \\
		\midrule
			NEES & 4.05 & 1.28 & 1.49 &  \textbf{1.07} \\
			RMSE robot (m) & 1.76 & 1.20 & 1.27 &  \textbf{1.18 }\\
			distance to iSAM (m) & 0.45 & 0.36 & 0.41 &  \textbf{0.30} \\
		Execution time (s) & \textbf{275} & 290 & 414 &  278 \\
		\bottomrule
	\end{tabular}
	\caption{Average  performances and computational time execution of the filters over the 1000 Monte-Carlo runs.  \label{fig:time_simu}} 
\end{figure}
\renewcommand{\figurename}{Fig.}
\setcounter{figure}{3} 

\section{Experimental Results}\label{sec:exp}

This section validates the proposed filter on multi-robot SLAM on the UTIAS dataset  \cite{leung_utias_2011}, to prove the feasibility and the benefits of the approach. Since both robocentric and OC-EKF are not straightforwardly applied to multi-robot SLAM, we compare our approach  to iSAM and a standard EKF only, both filter using a centralized scheme, although both can be used in decentralized estimation \cite{leung_decentralized_2012} (an OC-EKF has been derived only for cooperative localization  in \cite{huang_observability-based_2011}). 

The 2D indoor UTIAS dataset \cite{leung_utias_2011}  consists of a collection of 9 individual datasets of  20-70 min containing odometry and range and bearing measurements data from 5 robots, as well as  ground-truth for all robot poses and 15 landmark positions, and has been especially realized for studding the multi-robot SLAM problem. The forward velocity and angular velocity commands are logged at \SI{67}{Hz} as odometry data. The maximum forward velocity of a robot is \SI{0.16}{m/s}, and the maximum angular velocity is {0.35}{rad/s}. Each robot and landmark has a unique identification barcode, that are detected in rectified images captured by the camera on each robot. The encoded identification barcode as well as the range and bearing to each barcode is then extracted, where the camera on each robot is placed to align with the robot body frame.  The ground-truth pose  for each robot and ground-truth position for each landmark is provided by a Vicon motion capture system  at \SI{100}{Hz} with accuracy on the order of  \SI{e-3}{m}.

\subsection{Alternative Error Derivation}
Let $M$ robots navigate in an unknown environment of $P$ landmarks, where $(\bfp_R^i, \theta_i)$ is the pose of the $i$-th robot. Inspired from Section \ref{sec:extmul}, we suggest to treat each robot as a system with its own global orientation. Regarding landmarks,   we chose to consider each landmark as a system \eqref{eq:defx}, whose orientation is fixed and is associated  with the orientation $\theta^{j}$ of the robot that observes   this landmark for the first time (we have then $\theta^j_{n+1} = \theta^j_n$). Our error leads to   an increase of the dimension of the covariance matrix  $\bfP^e$, but clearly improves the filter accuracy, as shown below.

To recap the difference with EKF for the present problem:
\begin{itemize}
	\item The standard EKF conveys an estimate of the dispersion of the linear error, which is $(\hat\theta_i-\theta_i ,\hbfp_{R,i}-\bfp_{R,i})$ for the $i$-th robot and $\hbfp_{L,j}-\bfp_{L,j}
	$ for the $j$-th landmark.
	\item The proposed EKF conveys an estimate   of the dispersion of an alternative error defined as $(\hat\theta_i-\theta_i , \hbfp_{R,i}- \hbfR_i\bfR_i^T \bfp_{R,i})$ for the $i$-th robot and $(\hat\theta^j-\theta^j , \hbfp_{L,j}- \hbfR^j\bfR^{jT} \bfp_{L,j})$ for the $j$-th landmark.
\end{itemize}
This is an application of the method of Section \ref{sec:extmul}, where  each landmark position is associated to a fixed  orientation $\theta^{j}$. Note that,  deriving a robocentric EKF seems  non-trivial.

\begin{figure}
	\centering
	\includegraphics[width=10cm]{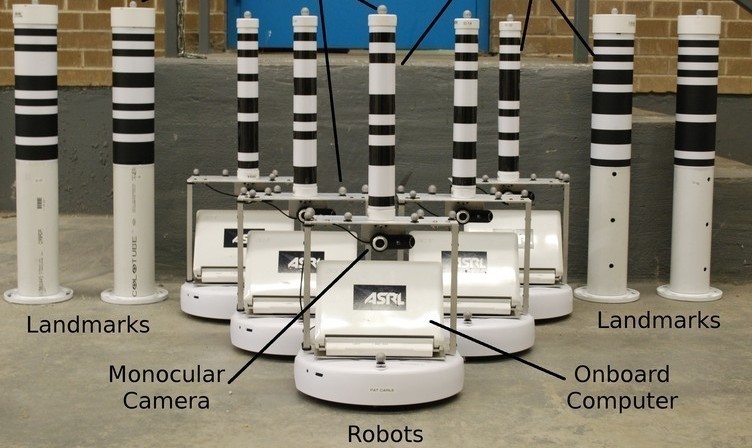}
	\caption{Pictures of the robots and landmarks used in UTIAS dataset \cite{leung_utias_2011}.\label{fig:photo}}
\end{figure}

\begin{figure}
	\centering
	\begin{tikzpicture}
\begin{axis}[width=10.5cm, height=5.5cm,
legend style={legend columns=-1, font=\small},
legend pos= north west,
bar width=1pt,
ylabel={robot position RMSE (m)},
ylabel style = {font=\footnotesize},
xtick=data,
xticklabels from table={\rmseExp}{filter},
xticklabel style = {font=\small},
xlabel={experiment number},
xlabel style={xshift=0ex,yshift=0.1cm},
ybar,
ybar interval=0.6,
xlabel style={xshift=-0ex,yshift=0.1cm},
enlargelimits=false,
ymin=0,ymax=0.25,
ytick={0,0.1,0.2,0.3},
yticklabel style={
	/pgf/number format/fixed,
	/pgf/number format/precision=2
}]
\addplot [color=blue,fill] table [x expr=\coordindex, y={standardEKF}] \rmseExp;
\addplot [color=green,fill] table [x expr=\coordindex, y={proposedEKF}] \rmseExp;
\addplot [color=yellow,fill] table [x expr=\coordindex, y={isam}] \rmseExp;
\legend{standard EKF, \textbf{proposed EKF}, iSAM}
\end{axis}

\end{tikzpicture}
	\caption{Evaluation of the accuracy performances in terms of  robot position RMSE.
		As in previous simulations,  the proposed EKF  systematically outperforms the standard EKF  and achieves comparable  results to iSAM.\label{fig:rmse_exp}}
\end{figure}
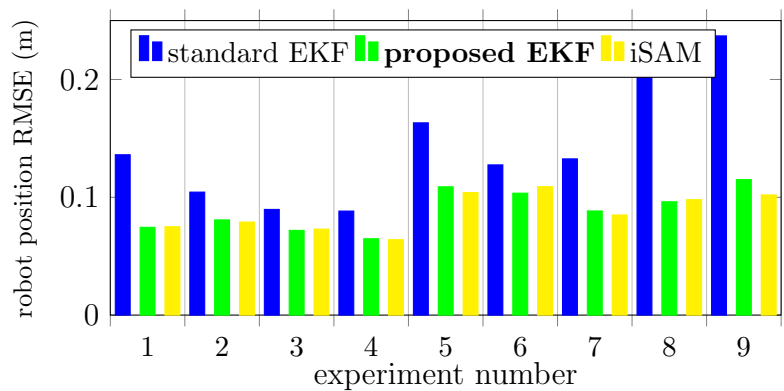

\renewcommand{\figurename}{Table}
\setcounter{figure}{1} 
\begin{figure}
	\centering
	\begin{tabular}{c||c|c|c}
		\toprule
		 & \small{standard EKF} & \small{\textbf{proposed EKF}} & \small{iSAM} \\
		\midrule
		\small{RMSE robots (m)} & 0.14 & \textbf{0.09} & \textbf{0.09} \\
		\bottomrule
	\end{tabular}
	\caption{Average RMSE over all datasets for the three methods.\label{fig:rmse_exp_tab}} 
\end{figure}
\renewcommand{\figurename}{Fig.}
\setcounter{figure}{5} 

\subsection{Experimental Results}
We conducted preliminary  tests to calibrate the motion model  and to characterize the noise properties of the motion and measurement models as follows: standard deviation on odometry  as 20 \% of the robot velocity, and standard deviation of range and bearing measurement as, respectively, \SI{0.5}{m} and \SI{3}{^\circ}. We then define a maximum observation range of \SI{5}{m}. This  corresponds to the most favorable setting for standard EKF.

Since the NEES is a statistical indicator and is very sensitive to parameter tuning, we focus on the RMSE, which is plotted on Figure \ref{fig:rmse_exp}   for the robot positions on all the available experiments. The RMSE for the standard EKF is summarized over all datasets in Figure \ref{fig:rmse_exp_tab}, with improvement for iSAM and the proposed EKF  of  as much as 50 \% compared to the standard EKF in the more challenging dataset 9 (where visual barriers reduce the number of barcode detections).

\section{Conclusion} \label{sec:con}

This work evidences  the EKF  for robot navigation is
not inherently inconsistent 
but the choice of the estimation error for linearization
 is pivotal: properly defining the error the EKF shall linearize yields consistency.  For SLAM, Monte-Carlo simulations and real  experiments have evidenced   our alternative EKF outperforms the EKF and achieves similar performance as state of the art iSAM. It thus offers an alternative to OC-EKF based on a sound mathematical theory anchored in geometry. Moreover the general theory goes beyond  basic  SLAM.   
 Future works concern the application of the method in various navigation problems and its derivation for both unscented Kalman filter and optimization techniques \cite{kaess_isam2:_2012, kummerle_g2o:_2011}. 
\section*{Acknowledgments}
We thank Guoquan Paul Huang of University of Delaware  for sharing his code of OC-EKF.

\appendix

\section{2D Mono-Robot Wheeled-SLAM}\label{sec:2dmono}
We detail  in this section the proposed filter for the 2D mono-robot wheeled-SLAM problem, which correspond to Section IV. This section starts by recalling the considering problem, details the proposed EKF and finishes with the standard EKF algorithm for the reader to compare and see what the differences are.

We consider a 2D SLAM system with one robot and $K$ landmarks. Let $\bfp_{R,n} \in \bbR^2$ be the position of the robot, $\bfR_n \in SO(2)$ the orientation of the robot, and $\bfp_{L,n}^j \in \bbR^2$ the position of the $j$-th landmark. The state  is  given as
\begin{align}
\bfx_n = \left(
\bfR_n,~ \bfp_{R,n},~ \bfp_{L,n}^1, ~\cdots,~\bfp_{L,n}^K
\right) \in \calX = SO(2) \times \bbR^{2(K+1)}.
\end{align}
The dynamics write 
\begin{align}
\bfx_{n+1}  &= f(\bfx_n,~\bfu_n,~\bfw_n) \\
 &=(\bfR_n \bfR(\omega_n +w_n^{\omega}),~  \bfp_{R,n} +\bfR_n \left( \bar \bfp_n + \bfw_n^{\bfp}\right),~ \bfp_{L,n}^1,~\cdots,~ \bfp_{L,n}^K),
\end{align}
 where 
\begin{align}
\bfu_n = \begin{bmatrix}
\omega_n \\
\bar \bfp_n
\end{bmatrix} \in  \bbR^{3}
\end{align}
denotes orientation and position increments typically measured through odometry, and $\bfR(\theta)$ is the rotation matrix of angle $\theta$. The noise in the propagation model is given as
\begin{align}
\bfw_n &= \begin{bmatrix} w_n^{\omega} \\ \bfw_n^{\bfp}\end{bmatrix} \in  \bbR^{3}, \\
\bfw_n &\sim \calN\left(\bfzero, \bfQ_n\right),
\end{align}
and contains noise on both the angular and position increments. 

The observation of the landmarks in the robot's frame writes
\begin{align}
\bfy_n &= \begin{bmatrix}
\bfy_n^1 \\
\vdots \\
\bfy_n^K
\end{bmatrix} \in \bbR^{2K}, \label{eq:y2d}\\
\bfy_n^k &= \tilde h(\bfR_n^T(\bfp_{L,n}^k-\bfp_{R,n})) + \bfn_n^k, ~k=1,\ldots,K,
\end{align}
where   for  any  landmark we have 
\begin{align}
\tilde h(\bfp) = \tilde h \left(\begin{bmatrix}
p_1 \\ p_2
\end{bmatrix} \right) = \begin{bmatrix}
\sqrt{p_1^2 + p_2^2} \\
\arctan 2 \left( p_2, p_1 \right)
\end{bmatrix}
\end{align}
which represents a range and bearing observation and $\bfn_n^k \sim \calN\left(\bfzero, \bfN_n^k\right)$ is the noise in the measurement of the $k$-th landmark, letting
\begin{align}
\bfN_n = \diag \left(\bfN_n^1, \cdots, \bfN_n^K\right) \in \bbR^{2K\times 2K}
\end{align}
be the covariance matrix for all the noise in the observation \eqref{eq:y2d}. Only a small fraction of the landmarks are observed at each step, i.e. only a subset of \eqref{eq:y2d} is used. We now detail the proposed EKF for the considered SLAM problem. This EKF first appears in \cite{barrau_ekf-slam_2015} and was then shown to remedy consistency issues in this context. 

\subsection{Proposed EKF Derivation}
For the considered problem, the non-linear error is defined as, see (25) 
\begin{equation}
\begin{aligned}
\eta(\bfx_n,\hbfx_n) &= (\bfR_n \hbfR_n^T, \hbfp_{R,n}- \hbfR_n\bfR_n^T \bfp_{R,n} ,  \cdots \\ & \quad \hbfp_{L,n}^1- \hbfR_n\bfR_n^T \bfp_{L,n}^1, \cdots, \hbfp_{L,n}^K- \hbfR_n\bfR_n^T \bfp_{L,n}^K ). \label{eq:eta}
\end{aligned}
\end{equation}
To linearize, we have the following first order vector approximation, see (26)-(27) 
\begin{align}
\check{\eta}(\bfx_n, \hbfx_n) &=  (\bfe_{R,n},~\hbfp_{R,n}- (\bfe_{R,n})_\times \bfp_{R,n},\cdots~ \nonumber
\\&\quad \hbfp_{L,n}^1- (\bfe_{R,n})_\times \bfp_{L,n}^1 ,\ldots,~\hbfp_{L,n}^K- (\bfe_{R,n})_\times \bfp_{L,n}^K  ) \in \bbR^{4+2K},\\
\bfR_n \hbfR_n^T &= \exp_{SO(2)}(\bfe_{R,n})\simeq \bfI + (\bfe_{R,n})_\times + o(\|\bfe_{R,n}\|^2).
\end{align}
The proposed filter operates in two steps: propagation and update, see Algorithm \ref{alg:ekf}. We now detail these two steps.

\begin{algorithm}[h]
	\KwIn{initial estimate $\hbfx_0$ and uncertainty matrix $\bfP_0^e$}
	\While{filter is running}{
		\SetKwBlock{Propagation}{Propagation}{end}
		\Propagation{
			\nl $\hbfx_{n|n-1} = f\left(\hbfx_{n-1|n-1},\bfu_n,\bfzero\right)$\;
			\nl $\bfP^e_{n|n-1} = \hat \bfF_n^e \bfP_{n-1|n-1}^e  (\hat \bfF_n^e)^T + \hat \bfG_n^e \bfQ_n (\hat \bfG_n^e)^T$\;
		}
		\SetKwBlock{Update}{Update}{end}
		\Update{
			
			\nl $\bfK_n = \hat \bfH_n^e \bfP_{n|n-1}^e \allowbreak / \left(\hat \bfH_n^e \bfP_{n|n-1}^e (\hat \bfH_n^e)^T + \hat \bfJ_n^e \bfR_n  (\hat \bfJ_n^e)^T\right)$\;
			\nl $\bfe_n^+ = \bfK_n \left( \bfy_n - h\left(\hbfx_{n|n-1},\bfzero\right)\right)$\;
			\nl $\hbfx_{n|n} = \psi\left(\hbfx_{n|n-1},\bfe_n^+\right)
			$; \tcp{state update}
			\nl $\bfP_{n|n}^e = \left(\bfI-\bfK_n\hat{\bfH}_n^e\right) \bfP_{n|n-1}^e$\;
		}		
	}
	\caption{EKF based on a non-linear state error \label{alg:ekf}}
\end{algorithm}

\subsection{Propagation}
At this step, we first propagate the state with the noise free model to compute $\hbfx_{n|n-1}$. We then propagate the covariance, where Jacobian are obtained after conserving only the first order error term in $\eta(\bfx_{n|n-1},\bfx_n)$. The Jacobian of the propagation are given as 
\begin{align}
\hat{\bfF}_n^e &=  \bfI, 
\end{align}
\begin{align}
\hat{\bfG}_n^e &= \begin{bmatrix}
1 & \bfzero  \\
(\hbfp_{R,n-1|n-1})_\times & \bfI  \\
(\hbfp_{L,n-1|n-1}^1)_\times & \bfzero \\
\vdots & \bfzero \\
(\hbfp_{L,n-1|n-1}^K)_\times & \bfzero
\end{bmatrix}.
\end{align}

\subsection{Update}
This step considers the observations of the landmarks. The Jacobian for for the measurements are given as
\begin{align}
&\hat\bfH_n^e  = \begin{bmatrix}
\hat\bfH_{n}^{e,1} \\ \vdots \\ \hat\bfH_{n}^{e,K}
\end{bmatrix}, \\
&\hat\bfH_{n}^{e,k} =
  \Delta h \left(\bfy_n^k\right) \bold{M}, \label{eq:hek} \\
&\Delta h \left(\bfy\right)  =  \begin{bmatrix} \frac{\bfy^T\bfJ^T}{\|\bfy\|^2} \\
\frac{\bfy^T}{\|\bfy\|} 
\end{bmatrix},
\end{align}where $\bold{M}$ is the matrix
\begin{align}
\bold{M} = \begin{bmatrix}
\bfzero  & \underbrace{-\hbfR_{n|n-1}^T}_{\text{columns 3 and 4}} & \bfzero & \cdots &  \underbrace{\hbfR_{n|n-1}^T}_{\text{columns 3+2$k$ and 4+2$k$}} & \bfzero & \cdots & \bfzero
\end{bmatrix}
\end{align}
and $\hat\bfJ^e_n = \bfI$. The non-zero parts of $\hat\bfH_{n}^{e,k}$ in \eqref{eq:hek} correspond to the error on the robot position and on the $k$-th landmark. Once the Kalman gain $\bfK_n$ is computed, we compute the innovation $\bfe_n^+$ and them update the state. The retraction required to update  the state is given as the exponential of $SE_{1+K}(2)$, see Section IV-A and see also \cite{barrau_ekf-slam_2015}. We have thus
\begin{align}
\hbfx_{n|n} &= \psi \left(\hbfx_{n|n-1},\bfe_n^+\right), \\
&= (\delta \bfR^+\hbfR_{n|n-1}, \delta \bfR^+\hbfp_{R,n|n-1} + \delta \bfp_R^+, \cdots\nonumber\\
& \delta \bfR^+\hbfp_{L,n|n-1}^1 + \delta \bfp_{L}^{1+}, \cdots, ~~\delta \bfR^+\hbfp_{L,n|n-1}^K + \delta \bfp_{L}^{K+} ),
\end{align}
where
\begin{align}
\delta \bfR^+ &= \bfR\left(\bfe_\bfR^+\right), \\
\delta \bfp_R^+ &= \bfA \bfe_R^+,\\
\delta \bfp_L^{k+} &= \bfA \bfe_L^{k+},k=1,\ldots,K,\\
\bfA  &= \begin{bmatrix}
\frac{\sin(\delta \bfe_\bfR^+)}{\delta \bfe_\bfR^+} & -\frac{1-\cos(\delta \bfe_\bfR^+)}{\delta \bfe_\bfR^+} \\ \frac{1-\cos(\delta \bfe_\bfR^+)}{\delta \bfe_\bfR^+} & \frac{\sin(\delta \bfe_\bfR^+)}{\delta \bfe_\bfR^+}
\end{bmatrix}, \\
\bfe_n^+ &= \begin{bmatrix}
\bfe_\bfR^+ \\
\bfe_R^+ \\
\bfe_L^{1+} \\
\vdots \\
\bfe_L^{K+}
\end{bmatrix}. \label{eq:e}
\end{align}
We conclude this step by updating the matrix covariance, see step 6 of Algorithm 2.

\subsection{Standard EKF Algorithm}
We provide in this section the standard EKF algorithm. The standard EKF follows Algorithm \ref{alg:ekf} with a simple linear error that we denote using the superscript ${\mathrm{\textsc{std}}}$.

In 3D, the difference between two rotation matrices does not make any sense. It is thus customary to use the difference in the sense of group multiplication on $SO(3)$, and this is what is referred to as ``standard'' EKF in the following. Thus, the state error on which the EKF is built is as follows:
\begin{align}
\eta^{\mathrm{\textsc{std}}}(\bfx_n,\hbfx_n) &=  (\bfR_n \hbfR_n^T, \hbfp_{R,n}-\bfp_{R,n}, \cdots \nonumber\\&\quad \hbfp_{L,n}^1-\bfp_{L,n}^1, \cdots, \hbfp_{L,n}^K-\bfp_{L,n}^K  ). 
\end{align}
The following first order vector approximation writes
\begin{align}
&\check{\eta}^{\mathrm{\textsc{std}}}(\bfx_n, \hbfx_n) =  (\bfe_{R,n},~\hbfp_{R,n}-\bfp_{R,n},\cdots\nonumber \\&\quad ~\hbfp_{L,n}^1- \bfp_{L,n}^1 ,\ldots,~\hbfp_{L,n}^K-\bfp_{L,n}^K  ) \in \bbR^{4+2K},\\
\bfR_n \hbfR_n^T &= \exp_{SO(2)}(\bfe_{R,n})\simeq \bfI + (\bfe_{R,n})_\times + o(\|\bfe_{R,n}\|^2).
\end{align}
The Jacobians and the retraction are given as
\begin{align}
\hat{\bfF}_n^{\mathrm{\textsc{std}}} &=  \mleft[\begin{array}{cc|ccc}
1 & \bfzero &  & \multirow{2}{*}{$\bfzero$} &  \\
\hbfR_{n-1|n-1}\bfJ \bar{\bfp}_{n-1|n-1}^T & \bfI &  &  & \\\hline
\multirow{2}{*}{$\bfzero$} & \multirow{2}{*}{$\bfzero$} & & \multirow{2}{*}{$\bfI$}\\
&&   \\
\end{array}\mright], 
\end{align}
\begin{align}
\bfJ = \begin{bmatrix}
0 & -1 \\
1 & 0
\end{bmatrix},
\end{align}
\begin{align}
\hat{\bfG}_n^{\mathrm{\textsc{std}}} &= \begin{bmatrix}
1 & \bfzero  \\
\bfzero & \hbfR_{n-1|n-1}  \\
\bfzero & \bfzero \\
\vdots & \bfzero \\
\bfzero & \bfzero
\end{bmatrix},
\end{align}
\begin{align}
\hat\bfH_n^{\mathrm{\textsc{std}}} &= \begin{bmatrix}
\hat\bfH_{n}^{\mathrm{\textsc{std}},1} \\ \vdots \\ \hat\bfH_{n}^{\mathrm{\textsc{std}},K}
\end{bmatrix}, \\
\hat\bfH_{n}^{\mathrm{\textsc{std}},k} &= \Delta h \left(\bfy_n^k\right)\bold{M}, 
\end{align}
with  $\bold{M}$ the matrix
 \[\begin{bmatrix}
\bold{A} & \underbrace{-\hbfR_{n|n-1}^T}_{\text{columns 3 and 4}} & \bfzero & \cdots &  \underbrace{\hbfR_{n|n-1}^T}_{\text{columns 3+2$k$ and 4+2$k$}} & \bfzero & \cdots & \bfzero
\end{bmatrix}\]
where we let 
\begin{align}
\bold A=-\bfJ \hbfR_{n|n-1}^T \left(\bfp_{L,n|n-1}^k - \bfp_{R,n|n-1}  \right) 
\end{align}
and
\begin{align}
\hbfx_{n|n} &= \psi\left(\hbfx_{n|n-1},\bfe_n^+\right), \\
&=  (\delta \bfR^+\hbfR_{n|n-1}, \hbfp_{R,n|n-1} +  \bfe_R^+, \cdots\nonumber \\&\quad  \hbfp_{L,n|n-1}^1 +  \bfe_{L}^{1+}, \cdots, \hbfp_{L,n|n-1}^K +  \bfe_{L}^{K+} ),
\end{align}
where
\begin{align}
\delta \bfR^+ = \bfR\left(\bfe_\bfR^+\right)
\end{align}
and $\bfe_n^+$ has the same form \eqref{eq:e} as for the proposed EKF, and  $\hat\bfJ^{\mathrm{\textsc{std}}}_n = \bfI$.

\section{3D Mono-Robot Wheeled-SLAM}\label{sec:3dmono}
We detail  in this section the proposed filter for the 3D mono-robot wheeled-SLAM problem. This section starts by recalling for the reader to compare. The similarities with the 2D problem (see Section \ref{sec:2dmono}) are obvious.

We consider a 3D SLAM system with one robot and $K$ landmarks. Let $\bfp_{R,n} \in \bbR^3$ be the position of the robot, $\bfR_n \in SO(3)$ the orientation of the robot, and $\bfp_{L,n}^j \in \bbR^3$ the position of the $j$-th landmark. The state  is  given as
\begin{align}
\bfx_n = \left(
\bfR_n,~ \bfp_{R,1},~ \bfp_{L,n}^1, ~\cdots,~\bfp_{L,n}^K
\right) \in \calX = SO(3) \times \bbR^{3(K+1)}.
\end{align}
The dynamics write 
\begin{align}
\bfx_{n+1}  &= f(\bfx_n,~\bfu_n,~\bfw_n) \\
&=(\bfR_n \bfR(\boomega_n +\bfw_n^{\boomega}),~  \bfp_{R,n} +\bfR_n \left( \bar \bfp_n + \bfw_n^{\bfp}\right),~ \bfp_{L,n}^1,~\cdots,~ \bfp_{L,n}^K),
\end{align}
where 
\begin{align}
\bfu_n = \begin{bmatrix}
\boomega_n \\
\bar \bfp_n
\end{bmatrix} \in  \bbR^{6}
\end{align}
denotes orientation and position increments typically measured through odometry, and $\bfR(\boomega) = \exp_{SO(3)}\left(\boomega\right)$. The noise in the propagation model is given as
\begin{align}
\bfw_n &= \begin{bmatrix} \bfw_n^{\boomega} \\  \bfw_n^{\bfp} \end{bmatrix} \in  \bbR^{6}, \\
\bfw_n &\sim \calN\left(\bfzero, ~\bfQ_n\right),
\end{align}
and contains noise on both the angular and position increments. 

The observation of the landmarks in the robot's frame is given as
\begin{align}
\bfy_n &= \begin{bmatrix}
\bfy_n^1 \\
\vdots \\
\bfy_n^K
\end{bmatrix} \in \bbR^{3K}, \label{eq:y3d}\\
\bfy_n^k &= \tilde h(\bfR_n^T(\bfp_{L,n}^k-\bfp_{R,n})) + \bfn_n^k, ~k=1,\ldots,K,
\end{align}
where the observation model for one landmark
\begin{align}
\tilde h(\bfp) = \tilde h \left(\begin{bmatrix}
p_1 \\ p_2 \\ p_3
\end{bmatrix} \right) = \begin{bmatrix}
 p_1/p_3\\
p_2/p_3
\end{bmatrix}
\end{align}
represents a perspective projection observation given e.g. by a monocular camera and $\bfn_n^k \sim \calN\left(\bfzero, \bfN_n^k\right)$ is the noise in the measurement of the $k$-th landmark, letting
\begin{align}
\bfN_n = \diag \left(\bfN_n^1, \cdots, \bfN_n^K\right) \in \bbR^{3K \times 3K}
\end{align}
be the covariance matrix for all the noise in the observation \eqref{eq:y3d}. Only a small fraction of the landmarks are observed at each step, i.e. only a subset of \eqref{eq:y3d} is used. We now detail the proposed EKF for the considered SLAM problem.

\subsection{Proposed EKF Derivation}
For the considered problem, the non-linear error is defined as  in (31)  
\begin{align}
\eta(\bfx_n,\hbfx_n) &= \big(\bfR_n \hbfR_n^T, \hbfp_{R,n}- \hbfR_n\bfR_n^T \bfp_{R,n} , \hbfp_{L,n}^1- \hbfR_n\bfR_n^T \bfp_{L,n}^1, \cdots, \nonumber \\
& \hbfp_{L,n}^K- \hbfR_n\bfR_n^T \bfp_{L,n}^K \big). 
\end{align}
To linearize, we have the following first order vector approximation, see (32)-(33) 
\begin{align}
\check{\eta}(\bfx_n, \hbfx_n) &= \big(\bfe_{R,n},~\hbfp_{R,n}- (\bfe_{R,n})_\times \bfp_{R,n},~\hbfp_{L,n}^1- (\bfe_{R,n})_\times \bfp_{L,n}^1 ,\cdots, \nonumber\\
&~\hbfp_{L,n}^K- (\bfe_{R,n})_\times \bfp_{L,n}^K  \big) \in \bbR^{6+3K},\\
\bfR_n \hbfR_n^T &= \exp_{SO(3)}(\bfe_{R,n})\simeq \bfI + (\bfe_{R,n})_\times + o(\|\bfe_{R,n}\|^2).
\end{align}
The proposed filter operates in two steps: propagation and update, see Algorithm \ref{alg:ekf}. We now detail these two steps.

\subsection{Propagation}
At this step, we first propagate the state with the noise free model to compute $\hbfx_{n|n-1}$. We then propagate the covariance, where Jacobian are obtained after conserving only the first order error term in $\eta(\bfx_{n|n-1},\bfx_n)$. The Jacobians of the propagation are given as 
\begin{align}
\hat{\bfF}_n^e &=  \bfI, 
\end{align}
\begin{align}
\hat{\bfG}_n^e &= \begin{bmatrix}
1 & \bfzero  \\
(\hbfp_{R,n-1|n-1})_\times & \bfI  \\
(\hbfp_{L,n-1|n-1}^1)_\times & \bfzero \\
\vdots & \bfzero \\
(\hbfp_{L,n-1|n-1}^K)_\times & \bfzero
\end{bmatrix}.
\end{align}

\subsection{Update}
This step considers the observations of the landmarks. The Jacobian for for the measurements are given as
\begin{align}
\hat\bfH_n^e &= \begin{bmatrix}
\hat\bfH_{n}^{e,1} \\ \vdots \\ \hat\bfH_{n}^{e,K}
\end{bmatrix}, \\
\hat\bfH_{n}^{e,k} &= \Delta h \left(\bfy_n^k\right)
\begin{bmatrix}
\bfzero  & \underbrace{-\hbfR_{n|n-1}^T}_{\text{columns 3 and 4}} & \bfzero & \cdots &  \underbrace{\hbfR_{n|n-1}^T}_{\text{columns 3+2$k$ and 4+2$k$}} & \bfzero & \cdots & \bfzero
\end{bmatrix}, \\
\Delta h \left(\bfy\right) &= \begin{bmatrix}
1 & 0 & -y_1/y_3^2 \\
0 & 1 & -y_2/y_3^2
\end{bmatrix},
\end{align}
and $\hat\bfJ^e_n = \bfI$. Once the Kalman gain $\bfK_n$ is computed, we compute the innovation $\bfe_n^+$ and them update the state. The retraction required for updated the state is given as the exponential of $SE_{1+K}(3)$, see Section IV-A. We have thus
\begin{align}
\hbfx_{n|n} &= \psi\left(\hbfx_{n|n-1},\bfe_n^+\right), \\
&= \big(\delta \bfR^+\hbfR_{n|n-1}, \delta \bfR^+\hbfp_{R,n|n-1} + \delta \bfp_R^+,  \delta \bfR^+\hbfp_{L,n|n-1}^1 + \delta \bfp_{L}^{1+}, \cdots, \nonumber \\
& ~\delta \bfR^+\hbfp_{L,n|n-1}^K + \delta \bfp_{L}^{K+}\big),
\end{align}
where
\begin{align}
\bfe_n^+ = \begin{bmatrix}
 \bfe_R^+ \\
 \bfe_R^+ \\
 \bfe_L^{1+} \\
\vdots \\
 \bfe_L^{K+}
\end{bmatrix},
\end{align}
\begin{align}
\left[\begin{array}{c|cccc}\arraycolsep=4pt
\delta \bfR^+ & \delta \bfp_R^+ & \bfp_L^{1+} &\cdots & \delta \bfp_L^{K+} \\
\hline
\bfzero &\multicolumn{4}{c}{\bfI}\end{array}\right] = \nonumber \\
\bfI + \bfS + \frac{1-\cos(\|\bfe_R^+\|)}{\|\bfe_R^+\|^2}\bfS^2 + \frac{\|\bfe_R^+\|-\sin(\|\bfe_R^+\|)}{\|\bfe_R^+\|^3}\bfS^3 
\end{align}
\begin{align}
\bfS = 	\left[
\begin{array}{c|cccc} \arraycolsep=1.4pt
(\bfe_R^+)_\times & \bfe_{R}^+ & \bfe_{L}^{1+} &\cdots & \bfe_{L}^{K+}\\
\hline
\bfzero &\multicolumn{4}{c}{\bfzero} 
\end{array}\right].
\end{align}
We conclude this step by updating the matrix covariance.

\subsection{Standard EKF Algorithm}
We provide in this section the standard EKF algorithm. The standard EKF follows Algorithm \ref{alg:ekf} with a different error as the proposed and where we use the superscript ${\mathrm{\textsc{std}}}$.

The non-linear error is defined as
\begin{align}
\eta^{\mathrm{\textsc{std}}}(\bfx_n,\hbfx_n) = \left(\bfR_n \hbfR_n^T, \hbfp_{R,n}-\bfp_{R,n}, \hbfp_{L,n}^1-\bfp_{L,n}^1, \cdots, \hbfp_{L,n}^K-\bfp_{L,n}^K \right). 
\end{align}
The following first order vector approximation writes
\begin{align}
\check{\eta}^{\mathrm{\textsc{std}}}(\bfx_n, \hbfx_n) &= \left(\bfe_{R,n},~\hbfp_{R,n}-\bfp_{R,n},~\hbfp_{L,n}^1- \bfp_{L,n}^1 ,\ldots,~\hbfp_{L,n}^K-\bfp_{L,n}^K  \right) \in \bbR^{4+2K},\\
\bfR_n \hbfR_n^T &= \exp_{SO(3)}(\bfe_{R,n})\simeq \bfI + (\bfe_{R,n})_\times + o(\|\bfe_{R,n}\|^2).
\end{align}
The Jacobians and the retraction are given as
\begin{align}
\hat{\bfF}_n^{\mathrm{\textsc{std}}} &=  \mleft[\begin{array}{cc|ccc}
\bfI & \bfzero &  & \multirow{2}{*}{$\bfzero$} &  \\
\hbfR_{n-1|n-1} \left( \bar{\bfp}_{n-1|n-1}\right)_{\times} & \bfI &  &  & \\\hline
\multirow{2}{*}{$\bfzero$} & \multirow{2}{*}{$\bfzero$} & & \multirow{2}{*}{$\bfI$}\\
&&   \\
\end{array}\mright], 
\end{align}
\begin{align}
\hat{\bfG}_n^{\mathrm{\textsc{std}}} &= \begin{bmatrix}
\bfI & \bfzero  \\
\bfzero & \hbfR_{n-1|n-1}  \\
\bfzero & \bfzero \\
\vdots & \bfzero \\
\bfzero & \bfzero
\end{bmatrix},
\end{align}
\begin{align}
\hat\bfH_n^{\mathrm{\textsc{std}}} &= \begin{bmatrix}
\hat\bfH_{n}^{\mathrm{\textsc{std}},1} \\ \vdots \\ \hat\bfH_{n}^{\mathrm{\textsc{std}},K}
\end{bmatrix},
\end{align}
\begin{align}
\hat\bfH_{n}^{\mathrm{\textsc{std}},k} &= \Delta h \left(\bfy_n^k\right)
\Big[-\hbfR_{n|n-1}^T \left(\bfp_{L,n|n-1}^k - \bfp_{R,n|n-1}  \right)_\times  ~~~ \underbrace{-\hbfR_{n|n-1}^T}_{\text{columns 3 and 4}} \nonumber\\
&\bfzero ~~ \cdots ~~  \underbrace{\hbfR_{n|n-1}^T}_{\text{columns 3+2$k$ and 4+2$k$}} ~~ \bfzero ~~ \cdots ~~ \bfzero
\Big],
\end{align}

\begin{align}
\hbfx_{n|n} &= \psi\left(\hbfx_{n|n-1},\bfe_n^+\right), \\
&= \big(\delta \bfR^+\hbfR_{n|n-1}, \hbfp_{R,n|n-1} + \delta \bfp_R^+,  \hbfp_{L,n|n-1}^1 + \delta \bfp_{L,n|n-1}^{1+}, \cdots, \nonumber\\
& \hbfp_{L,n|n-1}^K + \delta \bfp_{L,n|n-1}^{K+}\big),
\end{align}
where
\begin{align}
\delta \bfR^+ = \exp_{SO(3)}\left(\bfe_\bfR^+\right),
\end{align}
and $\bfe_n^+$ has the same form \eqref{eq:e} as for the proposed EKF, and  $\hat\bfJ^{\mathrm{\textsc{std}}}_n = \bfI$.

\section{2D Multi-Robot Wheeled-SLAM}\label{sec:2dmulti}
We detail  in this section the proposed filter for the 2D multi-robot wheeled-SLAM problem. This section starts by recalling the considering problem and details the proposed EKF. The similarities with the mono-robot problem (see Section \ref{sec:2dmono}) are immediate.

We consider a 2D SLAM system with $M$ robots and $K$ landmarks. Let $\bfp_{R,n}^m \in \bbR^2$ be the position of the $m$-th robot, $\bfR_n^m \in SO(2)$ the orientation of the $m$-th robot, and $\bfp_{L,n}^j \in \bbR^2$ the position of the $j$-th landmark. The state  is  given as
\begin{align}
\bfx_n &= \left(
\bfR_n^1, \ldots, \bfR_n^M,~ \bfp_{R,n}^1, \ldots, \bfp_{R,n}^{M},~ \bfp_{L,n}^1, ~\cdots,~\bfp_{L,n}^K
\right) \\
&\in \calX = SO(2) \times \bbR^{2(K+M)}. \nonumber
\end{align}
The dynamics writes
\begin{align}
\bfx_{n+1} &= f(\bfx_n,~\bfu_n,~\bfw_n) \\
&=(\bfR_n^1 \bfR(\omega_n^1 +w_n^{\omega, 1}),\ldots, \bfR_n^M \bfR(\omega_n^M +w_n^{\omega, M}),~  \bfp_{R,n}^1 +\bfR_n^1 \left( \bar \bfp_n^1 + \bfw_n^{\bfp, 1}\right), \nonumber \\ 
&\ldots,~  \bfp_{R,n}^M +\bfR_n^M \left( \bar \bfp_n^M + \bfw_n^{\bfp, M}\right), \bfp_{L,n}^1,~\cdots,~ \bfp_{L,n}^K)
\end{align}
where 
\begin{align}
\bfu_n &=  \begin{bmatrix} \bfu_n^1 \\
\vdots \\
\bfu_n^M  \end{bmatrix} \in  \bbR^{3M}, \\
\bfu_n^m &= \begin{bmatrix}
\omega_n^m \\
\bar \bfp_n^m
\end{bmatrix} \in  \bbR^3,
\end{align}
denotes orientation and position increments typically measured through odometry for each robot, and $\bfR(\theta)$ is the rotation matrix of angle $\theta$. The noise in the propagation model is given as
\begin{align}
\bfw_n &= \begin{bmatrix}
\bfw_n^1 \\ \vdots \\ \bfw_n^M 
\end{bmatrix} \in  \bbR^{3M} \\
\bfw_n^m &= \begin{bmatrix}w_n^{\omega, m} \\ \bfw_n^{\bfp, m} \end{bmatrix} \\
\bfw_n &\sim \calN\left(\bfzero, \bfQ_n\right),
\end{align}
and contains noise on both the angular and position increments. 

The observations of the landmarks robots in the $M$ robot's frames are given as
\begin{align}
\bfy_n &= \begin{bmatrix}
\bfy_n^{1,1} \\
\vdots \\
\bfy_n^{K+M-1, M}
\end{bmatrix} \in \bbR^{2M(K+M-1)}, \\
\bfy_n^{k, m} &= \tilde h(\bfR_n^{mT}(\bfp_{L,n}^k-\bfp_{R,n}^m)) + \bfn_n^{k,m}, ~k=1,\ldots,K,~ m=1,\ldots,M \\
\bfy_n^{k, m} &= \tilde h(\bfR_n^{mT}(\bfp_{R,n}^{k-K}-\bfp_{R,n}^m)) + \bfn_n^{k, m}, \\
&~k=K+1,\ldots,K+M,~ m=1,\ldots,M, ~ k \neq m \nonumber
\end{align}
where the observation model for one landmark or one robot
\begin{align}
\tilde h(\bfp) = \tilde h \left(\begin{bmatrix}
p_1 \\ p_2
\end{bmatrix} \right) = \begin{bmatrix}
\sqrt{p_1^2 + p_2^2} \\
\arctan 2 \left( p_2, p_1 \right)
\end{bmatrix}
\end{align}
represents a range and bearing observation and $\bfn_n^{k,m} \sim \calN\left(\bfzero, \bfN_n^{k,m}\right)$ is the noise in the measurement, letting
\begin{align}
\bfN_n = \diag \left(\bfN_n^{1,1}, \cdots, \bfN_n^{K+M-1, M}\right) \in \bbR^{2M(K+M-1)\times 2M(K+M-1)}
\end{align}
be the covariance matrix for all the noise in the observation \eqref{eq:y2d}. Only a small fraction of the landmarks are observed at each step, i.e. only a subset of \eqref{eq:y2d} is used. We now detail the proposed EKF for the considered SLAM problem.

\subsection{Proposed EKF Derivation}
For the considered problem, the non-linear error is defined as,  (25) 
\begin{align}
\eta(\bfx_n,\hbfx_n) &= \big(\bfR_n^1 \hbfR_n^{1T}, \ldots, \bfR_n^M \hbfR_n^{MT} \nonumber\\
&~~~~ \hbfp_{R,n}^1- \hbfR_n^1\bfR_n^{1T} \bfp_{R,n}^1 , \ldots  \hbfp_{R,n}^M- \hbfR_n^M\bfR_n^{MT} \bfp_{R,n}^M, \nonumber\\
&~~~~\hbfp_{L,n}^1- \hbfR_{L}^1\bfR_{L}^{1T} \bfp_{L,n}^1, \cdots, \hbfp_{L,n}^K- \hbfR_{L}^K\bfR_{L}^{KT} \bfp_{L,n}^K \big). 
\end{align}
To linearize, we have the following first order vector approximation, see  (26)-(27) 
\begin{align}
\check{\eta}(\bfx_n, \hbfx_n) &= \big(\bfe_{R,n}^1,\ldots, ~\bfe_{R,n}^M, \nonumber\\
&~\hbfp_{R,n}^1- (\bfe_{R,n}^1)_\times \bfp_{R,n}^1,\ldots, ~\hbfp_{R,n}^M- (\bfe_{R,n}^M)_\times \bfp_{R,n}^M \nonumber\\
&~\hbfp_{L,n}^1- (\bfe_{RL,n}^1)_\times \bfp_{L,n}^1 ,\ldots,~\hbfp_{L,n}^K- (\bfe_{RL,n}^K)_\times \bfp_{L,n}^K  \big) \in \bbR^{3(M+K)},\\
\bfR_n \hbfR_n^T &= \exp_{SO(2)}(\bfe_{R,n})\simeq \bfI + (\bfe_{R,n})_\times + o(\|\bfe_{R,n}\|^2).
\end{align}
The proposed filter operates in two steps: propagation and update, see Algorithm \ref{alg:ekf}. We now detail these two steps.

\subsection{Propagation}
TODO
During these state, we first propagate the state with the noise free model to compute $\hbfx_{n|n-1}$. We then propagate the covariance, where Jacobian are obtained after conserving only the first order error term in $\eta(\bfx_{n|n-1},\bfx_n)$. The Jacobian of the propagation are given as 
\begin{align}
\hat{\bfF}_n^e &=  \bfI, 
\end{align}
\begin{align}
\hat{\bfG}_n^e &= \begin{bmatrix}
\bfI & \bfzero  \\
(\hbfp_{R,n-1|n-1})_\times & \bfI  \\
(\hbfp_{L,n-1|n-1}^1)_\times & \bfzero \\
\vdots & \bfzero \\
(\hbfp_{L,n-1|n-1}^K)_\times & \bfzero
\end{bmatrix}.
\end{align}

\subsection{Update}
This step considers the observations of the landmarks. The Jacobian for for the measurements are given as
\begin{align}
\hat\bfH_n^e &= \begin{bmatrix}
\hat\bfH_{n}^{e,1} \\ \vdots \\ \hat\bfH_{n}^{e,K}
\end{bmatrix}, \\
\hat\bfH_{n}^{e,k} &=
\Delta h \left(\bfy_n^k\right) \begin{bmatrix}
\bfzero  & \underbrace{-\hbfR_{n|n-1}^T}_{\text{columns 3 and 4}} & \bfzero & \cdots &  \underbrace{\hbfR_{n|n-1}^T}_{\text{columns 3+2$k$ and 4+2$k$}} & \bfzero & \cdots & \bfzero
\end{bmatrix}, \\
\Delta h \left(\bfy\right) &=  \begin{bmatrix} \frac{\bfy^T\bfJ^T}{\|\bfy\|^2} \\
\frac{\bfy^T}{\|\bfy\|} 
\end{bmatrix}
\end{align}
and $\hat\bfJ^e_n = \bfI$. Once the Kalman gain $\bfK_n$ is computed, we compute the innovation $\bfe_n^+$ and them update the state. The retraction required for updated the state is given as the exponential of $SE_{1+K}(2)$, see Section IV-A of \cite{brossard2018exploiting}. We have thus
\begin{align}
\hbfx_{n|n} &= \psi\left(\hbfx_{n|n-1},\bfe_n^+\right), \\
&= \big(\delta \bfR^+\hbfR_{n|n-1}, \delta \bfR^+\hbfp_{R,n|n-1} + \delta \bfp_R^+,  \delta \bfR^+\hbfp_{L,n|n-1}^1 + \delta \bfp_{L}^{1+}, \cdots, \nonumber \\
&\delta \bfR^+\hbfp_{L,n|n-1}^K + \delta \bfp_{L}^{K+}\big)
\end{align}
where
\begin{align}
\delta \bfR^+ &= \bfR\left(\bfe_\bfR^+\right), \\
\delta \bfp_R^+ &= \bfA \bfe_R^+,\\
\delta \bfp_L^{k+} &= \bfA \bfe_L^{k+},k=1,\ldots,K,\\
\bfA  &= \begin{bmatrix}
\frac{\sin(\delta \bfe_\bfR^+)}{\delta \bfe_\bfR^+} & -\frac{1-\cos(\delta \bfe_\bfR^+)}{\delta \bfe_\bfR^+} \\ \frac{1-\cos(\delta \bfe_\bfR^+)}{\delta \bfe_\bfR^+} & \frac{\sin(\delta \bfe_\bfR^+)}{\delta \bfe_\bfR^+}
\end{bmatrix}, \\
\bfe_n^+ &= \begin{bmatrix}
\bfe_\bfR^+ \\
\bfe_R^+ \\
\bfe_L^{1+} \\
\vdots \\
\bfe_L^{K+}
\end{bmatrix}. 
\end{align}
We conclude this step by updating the matrix covariance.

\bibliographystyle{plain}
\bibliography{biblio}

\end{document}